\pdfoutput=1
\documentclass{vldb}

\usepackage{graphicx,array,multirow,url,subcaption,amssymb,fixltx2e,latexsym,amsmath,booktabs,color,tikz}
\usetikzlibrary{matrix,arrows,positioning}
\usepackage[ruled,linesnumbered,algonl,vlined,oldcommands]{algorithm2e}

\newcommand{\nop}[1]{}
\graphicspath{{images/}}

\begin{document}

\newcommand{\reminder}[1]{\textbf{[** #1 **]}}  


\newcommand{\todo}[1]{\vspace{5 mm}\par \noindent
 \marginpar
{\textsc{ToDo}}
 \framebox{\begin{minipage}[c]{0.95 \textwidth}
 \tt #1 \end{minipage}}\vspace{5 mm}\par}

\newcommand{\eat}[1]{}

\def\papernumber #1 raised #2 {
  \vspace{-#2}
  \vbox to 0pt{\framebox{\bf Paper Number: #1}}
}


\newcommand{\group}{g}
\newcommand{\groupw}{\group^w}
\newcommand{\minsup}{minsup}
\newcommand{\UniRank}{UniRank}
\newcommand{\ST}{PTWCorr}
\newcommand{\UniWCorr}{UniWCorr}
\newcommand{\DNG}{DeTeNG}
\newcommand{\UniCorr}{UniCorr}
\newcommand{\PODD}{GMoDD}
\newcommand{\gsim}{\Omega}
\newcommand{\tsize}{T}
\newcommand{\real}{\mathbb{R}}
\newcommand{\pair}{\mathcal{S}}
\newcommand{\SEPR}{\hspace*{0.6cm}}
\newcommand{\sepr}{\hspace*{0.4cm}}
\newcommand{\smsep}{\hspace*{0.2cm}}
\newcommand{\s}{\hspace*{2mm}}
\newcommand{\noparag}{\hspace*{-\parindent}}
\newcommand{\parag}{\hspace*{ \parindent}}
\newcommand{\smgap}{-5pt}
\newcommand{\para}[1]{\noindent\textbf{#1}}

\newcommand{\un}{\underline}
\newcommand{\pf}{\sf}                   
\newcommand{\fs}{\small}                
\newcommand{\schf}{\footnotesize\tt}    
\newcommand{\PFB}[1]{\mbox{\pf \footnotesize #1}}
\newcommand{\PFBs}[1]{\mbox{\pf \scriptsize #1}}
\renewcommand{\ttdefault}{pcr}

\newcommand{\SetOf}{\PFB{{SetOf}}}
\newcommand{\Rcd}{\PFB{{Rcd}}}
\newcommand{\Choice}{\PFB{{Choice}}}
\newcommand{\Str}{\PFB{{str}}}
\newcommand{\Int}{\PFB{{int}}}
\newcommand{\Flt}{\PFB{{float}}}

\newcommand{\Set}{\PFB{\un{set}}}
\newcommand{\Let}{\PFB{\un{let}}}
\newcommand{\mf}{\PFB{\un{MF}}}
\newcommand{\mr}{\PFB{\un{MR}}}
\newcommand{\In}{~\PFB{\un{in}}~}
\newcommand{\Where}{\PFB{\un{where}}}
\newcommand{\With}{\PFB{\un{with}}}
\newcommand{\Select}{\PFB{\un{select}}}
\newcommand{\Orderby}{\PFB{\un{order by}}}
\newcommand{\Groupby}{\PFB{\un{group by}}}
\newcommand{\Limit}{\PFB{\un{limit}}}
\newcommand{\From}{\PFB{\un{from}}}
\newcommand{\EAnd}{~\PFB{\un{and}}~}
\newcommand{\Tuple}{\PFB{~\un{struct}~}}
\newcommand{\Exists}{\PFB{\un{exists}}}
\newcommand{\Element}{\PFB{~\un{element}~}}
\newcommand{\Every}{\PFB{\un{every}}}
\newcommand{\Satisfies}{\PFB{\un{satisfies}}}

\newcommand{\Ins}{~\PFBs{\un{in}}~}
\newcommand{\Wheres}{\PFBs{\un{where}}}
\newcommand{\Withs}{\PFBs{\un{with}}}
\newcommand{\Selects}{\PFBs{\un{select}}}
\newcommand{\Fors}{\PFBs{\un{for}}}
\newcommand{\Returns}{\PFBs{\un{return}}}
\newcommand{\Froms}{\PFBs{\un{from}}}
\newcommand{\EAnds}{~\PFBs{\un{and}}~}
\newcommand{\Tuples}{\PFBs{~\un{struct}~}}
\newcommand{\Existss}{\PFBs{\un{exists}}}
\newcommand{\Elements}{\PFBs{~\un{element}~}}
\newcommand{\Foreachs}{\PFBs{\un{foreach}}}
\newcommand{\Everys}{\PFBs{\un{every}}}
\newcommand{\Ifs}{\PFBs{\un{if}}}
\newcommand{\Thens}{\PFBs{\un{then}}}
\newcommand{\Satisfiess}{\PFBs{\un{satisfies}}}
\newcommand{\expl}{\mbox {\tt Expl}}
\newcommand{\itemsim}{\mbox {\tt ItemSim}}
\newcommand{\usersim}{\mbox {\tt UserSim}}
\newcommand{\calu}{\mbox {$\cal U$}}
\newcommand{\cali}{\mbox {$\cal I$}}
\newcommand{\nw}{\mbox {\tt Network}}
\newcommand{\ddu}{\mbox {$DD_u$}}
\newcommand{\ddju}{\mbox {$DD^J_u$}}
\newcommand{\ddcu}{\mbox {$DD^C_u$}}
\newcommand{\socrel}{\mbox {\tt SocRel}}

\newcommand{\from}[2]{{\sc from} {\bf #1: #2}}

\newcommand{\ch}{$\rightarrow$}
\newcommand{\mt}[1]{$#1$}
\newcommand{\mc}[1]{$\mathcal{#1}$}
\newcommand{\mcs}[2]{$\mathcal{#1}_{#2}$}

\newtheorem{definition}{Definition}
\newtheorem{corollary}{Corollary}
\newtheorem{theorem}{Theorem}

\newtheorem{Ob}{Observation}

\newenvironment{query}{
\footnotesize
\vspace*{\smgap}
\begin{tabbing} }
{\end{tabbing}
\normalsize\small\normalsize
\vspace*{\smgap}
}

\newenvironment{querys}{
\scriptsize
\vspace*{\smgap}
\begin{tabbing} }
{\end{tabbing}
\normalsize\small\normalsize
\vspace*{\smgap}
}

\newcommand{\secref}[1]{Section~\ref{#1}} 
\newcommand{\figref}[1]{Figure~\ref{#1}} 
\newcommand{\tbref}[1]{Table~\ref{#1}} 
\newcommand{\cass}{Hypothesis~}
\newcommand{\cequ}{Equation~}
\newcommand{\hide}[1]{} 
\newcommand{\ie}{{\sl i.e.}}
\newcommand{\eg}{{\sl e.g.}}
\newcommand{\etc}{{\sl etc.}}
\newcommand{\etal}{{\sl et al.}}
\newcommand{\adhoc}{{\sl ad hoc}}
\newtheorem{property}{Property}
\newtheorem{example}{Example}

\newcommand{\xueqing}[1]{\textcolor{red}{#1}}
\renewcommand{\topfraction}{0.9}	
\renewcommand{\bottomfraction}{0.8}	
\renewcommand{\dbltopfraction}{0.90}
\renewcommand{\textfraction}{0.07}
\renewcommand{\floatpagefraction}{0.7}	
\renewcommand{\dblfloatpagefraction}{0.7}	

\title{Scalable and Robust Construction of Topical Hierarchies}

\numberofauthors{1} 
\author{
\alignauthor
Chi Wang, Xueqing Liu, Yanglei Song, Jiawei Han\\
       \affaddr{Department of Computer Science}\\
       \affaddr{University of Illinois at Urbana-Champaign}\\
       \affaddr{Urbana, IL, USA}\\
       \email{\{chiwang1, xliu93, ysong44, hanj\}@illinois.edu}       
}

\maketitle
\begin{abstract}
Automated generation of high-quality topical hierarchies for a text collection is a dream problem in knowledge engineering with many valuable applications. In this paper a scalable and robust algorithm is proposed for constructing a hierarchy of topics from a text collection. We divide and conquer the problem using a top-down recursive framework, based on a tensor orthogonal decomposition technique. We solve a critical challenge to perform scalable inference for our newly designed hierarchical topic model. Experiments with various real-world datasets illustrate its ability to generate robust, high-quality hierarchies efficiently. Our method reduces the time of construction by several orders of magnitude, and its robust feature renders it possible for users to interactively revise the hierarchy.

\end{abstract}




\section{Introduction}


Automated, hierarchical organization of the concepts in a textual database at different levels of granularity is an important problem in knowledge engineering with many valuable applications such as information summarization, search and online analytical processing (OLAP).  With vast amount of text data and dynamic change of users' need, it is too costly to rely on human experts to do manual annotation and provide ready-to-use topical hierarchies.  Thus it is critical to create a robust framework for automated construction of high-quality topical hierarchies from texts in different domains.

\smallskip
\noindent \textbf{Limitation of prior work.} A main body of existing work uses a bag-of-words topic modeling approach to model word occurrences in the documents. They infer the whole hierarchy all at once by inference methods such as Gibbs sampling.  Such methods have the following bottlenecks:
\begin{enumerate}
\parskip -0.2ex
\item \textbf{Scalability}. The inference methods such as Gibbs sampling\eat{ and variational inference} are expensive, requiring multiple passes of the data. The number of passes has no theoretical bound, and typically needs to be several hundreds or thousands.
  \item \textbf{Robustness}. The inference\eat{ or variational inference} algorithms do not produce a unique solution in nature. The variance of different algorithm runs can be very large especially when the hierarchy is deep. This prevents a user from revising the local structure of a hierarchy (\eg, changing the number of branches of one node).
      \eat{Moreover, if one wants to expand a hierarchy by one level, the entire hierarchy will change due to  }
  \item \textbf{Interpretability}. The unigram representation of topics has limited human interpretability, especially for very specific topics. But it is challenging to integrate multigrams into topic hierarchy modeling in a scalable and robust manner.
\end{enumerate}

\noindent \textbf{Insight.} The following ideas lead to our proposed solution.

\begin{figure*}
\small
\tikzstyle{old}=[circle,
                                    thick,
                                    draw=black!60,
                                    fill=black!20,
                                    inner sep = 1pt,
                                    minimum size=0.7cm]
\begin{minipage}{.35\linewidth}

 \begin{tikzpicture}[scale=.35,auto=left]

  \node (t0)[old][label=left:o] at (9,10) {$t0$};
  \node (t1)[old][label=left:o/1]  at (5,6)  {$t1$};
  \node (t2)[old] at (9, 6) {$t2$};
  \node (t3)[old] at (13, 6) {$t3$};
  \node (t4)[old] at (2, 2) {$t4$};
  \node (t5)[old] at (5, 2) [label=below:o/1/2]{$t5$};
  \node (t6)[old] at (8, 2) {$t6$};
  \node (t7)[old] at (11, 2) {$t7$};
  \node (t8)[old] at (14, 2) {$t8$};
  \node (t9)[old] at (17, 2) {$t9$};
  
    \path[->]
    (t0) edge (t1)
    (t0) edge (t2)
    (t0) edge (t3)
    (t1) edge (t4)
    (t1) edge (t5)
    (t3) edge (t8)
    (t3) edge (t9)
    (t2) edge (t6)
    (t2) edge (t7)
    ;

\end{tikzpicture}
 \end{minipage}
 \hfill
\begin{minipage}{.7\textwidth}
\renewcommand{\arraystretch}{1.4}
 \begin{tabular}{l|m{4.4cm}|m{5.0cm}}
\textbf{Topic} &
\textbf{Word distribution} & \textbf{Representative phrase}\\  \hline
t0(o) & data:0.01, learning:0.01 \ldots & database system, machine learning \ldots\\ \hline
t1(o/1) & database:0.05, system:0.01 \ldots & database system, management \ldots  \\ \hline
t2(o/2) & information:0.1, retrieval:0.05 \ldots & information retrieval, web search \ldots  \\ \hline
t3(o/3) & learning:0.11, classification:0.01 \ldots & learning, classification, feature selection \ldots  \\ \hline
t4(o/1/1) & query:0.12, processing:0.07 \ldots & query processing, query optimization \ldots \\ \hline
t5(o/1/2) & system:0.08, distributed:0.03 \ldots & distributed database, concurrency control \ldots \\
\end{tabular}
\end{minipage}

 \caption{An example of the topical hierarchy. Each topic can be denoted by the path from root topic to it}
 \label{fig:eg}
\end{figure*}

\smallskip
\noindent
1. \emph{Consider a strategy of top-down recursive construction of a topical hierarchy, instead of inferring a complex hierarchical model all at once}. For example, in Figure~\ref{fig:eg}, we first infer the topics $t1$ to $t3$ at the first level and then infer subtopics for each of them.
Thus the problem of uncovering a big tree can be divided into a set of subproblems: uncovering subtrees, and then conquered by solving each subproblem in the same manner. Then we can focus on solving a simpler (yet still challenging) problem: uncovering topics for one level, with good scalability, robustness and interpretability.

\smallskip
\noindent
2. \emph{Compress the original data by collecting important statistics from the documents}, \eg, word co-occurrences, in order to infer the topics efficiently and robustly.  For one advantage, the inference based on the compressed information avoids the expensive, numerous passes of the data.  For another advantage, the compression reduces certain degree of randomness in the data. By carefully choosing the statistics and the inference method, we can uncover the topics with theoretical guarantee. This insight is supported by some very recent finding in the theory study~\cite{anandkumar2012}, and we leverage it as a most important basis to develop our recursive approach and justify it. \eat{is removed of the algorithm runs,and performing deterministic inference based on these statistics. that are sufficient for robust topic hierarchy construction. }

\smallskip
\noindent
3. \emph{Enhance the topic representation using frequent multigrams, mined from the documents and placed in the topic hierarchy according to the inferred topics}. We do not consider multigrams when inferring topics, due to the scalability consideration. But we perform lightweight posterior estimation of the distribution over the topics for each frequent multigram.  Then we can sort the multigrams as topical phrases for each topic. For example, a database topic should contain `database system,' `query processing' \etc\ as top-ranked phrases.

\smallskip
\noindent
We systematically develop our solution based on these insights. The following summarizes our main contributions:
\begin{itemize}
\parskip -0.2ex
\item We propose a new hierarchical topic model, which \hide{is simpler than all existing hierarchical topic models but }supports divide-and-conquer inference as mentioned above. We provide theoretical justification of doing so.

\item We develop a scalable tensor-based recursive orthogonal decomposition (STROD) method to infer the model. It inherits the nice theoretical properties of the tensor orthogonal decomposition algorithm~\cite{anandkumar2012}, but has significantly better scalability. To the best of our knowledge, it is the first scalable and robust algorithm for topical hierarchy construction.

\item Our experiments demonstrate that our method can scale up to datasets that are orders of magnitude larger than the state-of-the-art, while generating quality topic hierarchy that is comprehensible to users.
\end{itemize}

\section{Problem Formulation}
\label{sec:prelim}

The input is a corpus of $D$ documents. Every document $d_i$ can be segmented into a sequence of $l_i$ tokens: $d_{i,j}, j=1,\dots,l_i$. For convenience we index all the unique words in this corpus using a vocabulary of $V$ words. And $d_{i,j}=x, x\in\{1,\dots,V\}$ means that the $j$-th token in document $d_i$ is the $x$-th word in the vocabulary. Throughout this paper we use `word $x$' to refer to the $x$-th word in the vocabulary.

Given a corpus, our goal is to construct a topical hierarchy. A topical hierarchy is a tree of topics, where each child topic is about a more specific theme within the parent topic. Statistically, a topic $t$ is characterized by a probability distribution over words $\phi_t$. $\phi_{t,x}=p(x|t)\in [0,1]$ is the probability of seeing the word $x$ in topic $t$, and $\sum_{x=1}^V \phi_{t,x}=1$. For example, in a topic about the database research area, the probability of seeing ``database", ``system" and ``query" is high, and the probability of seeing ``speech", ``handwriting" and ``animation" is low. This characterization is advantageous in statistical modeling of text, but is weak in human interpretability, due to two reasons. First, unigrams may be ambiguous, especially across specific topics. \hide{For example, the word ``query" can have high probability to be observed in both database topic and information retrieval topic. For example, the word ``document'' may have one sense in an information retrieval topic (such as, an item to be retrieved), but another sense in database topic (such as, an action to be taken). }Second, the probability $p(x|t)$ reflects the popularity of a word $x$ in the topic $t$, but not its discriminating power. For example, a general word ``algorithm" may have higher probability than a discriminative word ``locking" in the database topic.

For these reasons, we choose to enhance the topic representation with ranked phrases. Phrases reduce the ambiguity of unigrams. And the ranking should reflect both their popularity and discriminating power for a topic. For example, the database topic can be described as: \{``database systems", ``query processing", ``concurrency control", $\ldots$\}.  A phrase can appear in multiple topics, though it will have various ranks in them.

Formally, we define a topical hierarchy as follows.

\begin{definition}[Topical Hierarchy]
A topical hierarchy is defined as a tree $\mathcal{T}$ in which each node is a topic. Every non-leaf topic $t$ has $C_t$ child topics. Topic $t$ is characterized by a probability distribution over words $\phi_t$, and presented as an ordered list of phrases $\mathcal{P}_t=\{P_{t,1},P_{t,2},\dots\}$, where ${P}_{t,i}$ is the phrase ranked at $i$-th position for topic $t$.
\end{definition}

The number of child topics $C_t$ of each topic can be specified as input, or decided automatically by the construction method. In both cases, we assume it is bounded by a small number $K$, such as 10. This is for efficient browsing of the topics. 
\hide{For example, if one wants to locate one of 100 topics, it is tedious to check all the 100 topics. A hierarchical organization of these 100 topics, such as 10 by 10 (10 topics under the root, and 10 child topics for each of them), helps reducing the overhead. 
One just needs to inspect twice, each time going through 10 topics at most. }
The number $K$ is named the \emph{width} of the tree $\mathcal{T}$.

For convenience, we denote a topic using the top-down path from root to this topic. The root topic is denoted as $o$. 
Every non-root topic $t$ is denoted by $\pi_t/\chi_t$, where $\pi_t$ is the notation of its parent topic, and $\chi_t$ is the index of $t$ among its siblings. 
For example, the topic $t1$ in Figure~\ref{fig:eg} is denoted as $o/1$, and its child $t5$ is denoted as $o/1/2$. The \emph{level} $h_t$ of a topic $t$ is defined to be the number of `/' in its notation. So
root topic is in level 0, and $t5$ is in level 2. The \emph{height} $H$ of a tree is defined to be the maximal level over all the topics in the tree. Clearly, the total number $T$ of topics is upper bounded by $\frac{K^{H+1}-1}{K-1}$.

\subsection{Desired Property}
\label{sec:property}
The following are the desired properties for a topical hierarchy construction method.

\smallskip
\noindent
\textbf{1. Scalability.} The scale of the problem is determined by these variables: the number of documents $D$, the vocabulary size $V$, the total length of documents $L$, the total number of topics $T$, and the width of the topical hierarchy $K$. These variables are not independent. For example, the average length of documents $L/D$ should be larger than 1, and the number of documents $D$ is usually much larger than the vocabulary size $V$.  Typically, the number of tokens $L$ is the dominant factor.
\hide{They typically satisfy $DV>>L>D>V>>|\mathcal{T}|>>K\approx 10$ \xueqing{(I think it is not rigorous to make such assumption)}. }For scalability the algorithm must have sublinear complexity with respect to $L$. When the dataset is too large to fit in memory, an ideal algorithm should only perform a small constant number of passes of the data.

\smallskip
\noindent
\textbf{2. Robustness.} A construction algorithm is robust in the following senses.
\begin{property}[Robust Recovery]
Suppose the data is generated from certain topic word distributions exactly following a generative process, the recovery is robust if the exact distributions can be found when sufficient data are given.
\end{property}
For example, in Figure~\ref{fig:eg}, if sufficient data are generated from the topic word distributions as shown on the right hand side, an robust recovery algorithm should return these distributions rather than other distributions.

In certain scenarios, part of the constructed hierarchy needs to be revised to customize for users' need. For example, one may want to change the number of branches or height of a subtree.
\begin{property}[Robust Revision]
	The revision to a subtree $\mathcal{T}(t)$ rooted at topic $t$ is robust, if every topic $t'$ not in the subtree $\mathcal{T}(t)$ remains intact word distribution in the returned hierarchy.
\end{property}
In Figure~\ref{fig:eg}, if one wants to partition topic $t1$ into 3 subtopics instead of 2, but also wants to keep other parts of the tree intact, a robust algorithm should not change the output of topic $t2,t3$ and $t6$ to $t10$. This property assures that the local change to a large hierarchy doesn not alter the remainder of the tree.

\smallskip
\noindent
\textbf{3. Interpretability.}
There are two aspects for the interpretability of a topical hierarchy.
i) Topic coherence: one can interpret the meaning of an individual topic given the ranked words and phrases from it; and
ii) Parent-child relationship: one can interpret the meaning of the edges between a parent topic and its child topics given the ranked words and phrases from them.

\section{Related Work}\label{sec:relwork}
\hide
{\begin{table}
\centering
\caption{\label{tb:property}Scalability and robustness of known methods}
\small
\begin{tabular}{|l|l|m{1.8cm}|m{1.8cm}|}
\toprule
\textbf{Method} &
\textbf{\# Data Passes} & \textbf{Robust Recovery} & \textbf{Robust Revision} \\  \hline
nCRP & unbounded & Yes & No \\ \hline
PAM & unbounded & Yes & No \\ \hline
hPAM & unbounded & Yes & No \\ \hline
rCRP & unbounded & Yes & No \\ \hline
nCRF & unbounded & Yes & No \\ \hline
splitLDA & unbounded & N/A & Yes \\ \hline
CATHY & 1 & No & Yes \\
\bottomrule
\end{tabular}
\end{table}
}
Statistical topic modeling techniques model documents as mixtures of multiple topics, while every topic is modeled as a distribution over words. Two important models are PLSA (probabilistic latent semantic analysis)~\cite{Hofmann01} and its Bayesian extension LDA (latent Dirichlet allocation)~\cite{Blei03}. They model the generative processes of the words for all the documents in a corpus. To generate each word, a latent topic label is first chosen from a pool of `flat topics' with a multinomial distribution. And then a word is sampled according to this topic's word distribution. With these generative assumptions, the unknown word distribution of every topic can be inferred so as to best explain the observed word occurrences in the documents.

Hierarchical topic models follow the same generative spirit. Instead of generating from a pool of flat topics, these models assume an internal hierarchical structure of the topics. Different models use different generative processes to simulate this hierarchical structure: nested Chinese Restaurant Process~\cite{Griffiths04}, Pachinko Allocation~\cite{Li06}, Hierarchical Pachinko Allocation~\cite{Mimno07}, recursive Chinese Restaurant Process~\cite{Kim2012}, \hide{nested Hierarchical Dirichlet Process~\cite{paisley2012}, }and nested Chinese Resturant Franchise~\cite{ahmed2013}. When these models are applied to constructing a topical hierarchy, the entire hierarchy must be inferred all at once from the corpus. They do not have the robust revision property.

The main inference methods for these topic models can be divided into two categories: Gibbs sampling~\cite{griffiths2004finding} and variational inference~\cite{Blei03}. They are essentially based on the Maximum Likelihood (ML) principle (in the general sense): find the best parameters that maximize the joint probability specified by a model. There has been a substantial amount of work on speeding up LDA inference, \eg, by leveraging sparsity~\cite{porteous2008fast,yao2009efficient,hoffman2012sparse} and parallelization~\cite{newman2009distributed,smola2010architecture,zhai2012mr}, or online learning mechanism~\cite{ahmed2011online,hoffman2013stochastic,Foulds2013}. Few of these ideas have been adopted by the hierarchical topic model studies.

These inference methods have no theoretical guarantee of convergence within a bounded number of iterations, and are nondeterministic either due to the sampling or the random initialization. Recently, a new inference method for LDA has been proposed based on a \emph{method of moments}, rather than ML. It is found to have nice convergence properties in theory~\cite{anandkumar2012}. However, the practical issue of scalability has not been solved in the theoretical work.

All of the hierarchical topic models follow the bag-of-words assumption, while some other extensions of LDA have been developed to model sequential n-grams~\cite{Wallach06,Wang07,Lindsey12}. No one has integrated them in a hierarchical topic model. It is obvious that the scalability and robustness issues will become more challenging in an integrated model. A practical approach is to separate the topic modeling part and the phrase mining part. Blei and Lafferty~\cite{Blei2009} have proposed to use a statistical test to find topical phrases, which is time-consuming. A much less expensive heuristic is studied in our previous work~\cite{Danilevsky14} and shown to be effective.

Finally, we briefly review a few alternative approaches to constructing a topical hierarchy. Pujara and Skomoroch~\cite{pujara:nips12} proposed to first run LDA on the entire corpus, and then split the corpus heuristically according to the results and run LDA on each split corpus individually. It is a recursive approach without an integrated generative process, so the robust recovery property is not applicable. Recursive clustering is used to cluster documents~\cite{fung2003hierarchical}, queries~\cite{Liu+12}, keywords~\cite{Wang13infl} \etc, to construct hierarchies of different kinds. CATHY~\cite{Wang13infl} is a recursive topical phrase mining framework, where the phrase mining and the topic discovery are also separated for efficiency purpose. It uses a word co-occurrence network to compress the documents and performs topic discovery using an EM algorithm. However, it is designed for short, content-representative text and also suffers the scalability and robustness issues.

\section{Hierarchical Topic Modeling}\label{sec:model}

In this section, we model how the documents are generated when the topical hierarchy is given. Based on that we develop our solution in the next section. 
Our hierarchical topic model is a unigram-based model that supports recursive inference. We first explain the motivation behind the model in Section~\ref{sec:motivation}, and then present it in Section~\ref{sec:tlda}.

\subsection{Motivation}\label{sec:motivation}

First, we advocate using a top-down recursive framework to construct the hierarchy level by level, instead of all at once. Compared with a gigantic non-recursive construction method, a recursive framework has the following advantages: (i) it facilitates robust revision; (ii) the scalability and robust recovery challenges of the topical hierarchy can be reduced by the divide-and-conquer philosophy; and (iii) parent-child relationship naturally follows the recursion.
However, it may have the following disadvantages: (i) it misses some complex dependency across the topics mentioned within a document;
and (ii) the error in recovering a parent topic may propagate to a child topic, which implies the critical importance of robust recovery for top levels.
\hide{Since the goal of this study is to construct a human interpretable hierarchy from a large dataset in a scalable and robust manner, we argue that the advantages of a recursive framework overweigh its disadvantages. The complex dependency across topics other than the tree structure can make it more accurate at modeling each individual document. But }Since our major concern is to discover the recurring topics of the corpus rather than each single document, we are willing to use a simple tree structure for better scalability. Meanwhile, we need to carefully design our method to have good robustness and interpretability. \hide{Also, the more complex the dependency is, the harder it is to recover robustly. }

Second, for scalability consideration, we do not model the n-grams in the model, but discover topical phrases separately after the model inference. This strategy is shown to be effective in previous work~\cite{Blei2009,Wang13CATHY}. We discuss the phrase mining and ranking in Section~\ref{sec:phrase}.

\hide{Most existing hierarchical topic models do not support recursive inference. CATHY~\cite{Wang13CATHY} models the generation of word networks recursively, but the EM inference is neither robust nor scalable. So }Based on these considerations, we propose a new hierarchical topic model that allows for scalable, robust recursive inference.

\begin{table}
\centering
\caption{\label{tb:notations}Notations used in our model}
\small
\begin{tabular}{|r|m{6.5cm}|}
\toprule
\textbf{Symbol} & \textbf{Description} \\  \hline
$D$ & the number of documents in the corpus\\ \hline
$V$ & the number of unique words in the corpus \\ \hline
$d_{i,j}$ & the $j$-th word in the $i$-th document \\ \hline
$l_i$ & the length (number of tokens) of document $d_i$\\ \hline
$L$ & the total number of tokens in the corpus\hide{length of all the documents} $\sum_{i=1}^D l_i$\\ \hline
$\pi_t$ & the parent topic of topic $t$ \\ \hline
{$\chi_{t}$} & the suffix of topic $t$'s notation ($t=\pi_t/\chi_t$)\\ \hline
$C_t$ & the number of child topics of topic $t$ \\ \hline
$o$ & the root topic \\ \hline
$\phi_{t}$  & the multinomial distribution
 over words in topic $t$ \\ \hline
$\alpha_{t}$ & the Dirichlet hyperparameter vector of topic $t$ \\ \hline
$\theta_{i,t}$  & the distribution over child topics of $t$ for $d_i$\\ \hline
$T$ & the total number of topics in the hierarchy\\ \hline
$\tau$ & the number of leaf topics in the hierarchy \\ 
\bottomrule
\end{tabular}
\end{table}

\subsection{Latent Dirichlet Allocation with Topic Tree}\label{sec:tlda}
\begin{figure}
\centering
\tikzstyle{state}=[circle,
                                    thick,
                                    draw=black!60,
                                    inner sep = 1.2pt,
                                    minimum size=0.5cm]
                                   
\tikzstyle{doc}=[circle,
                                    thick,
                                    draw=black!60,
                                    fill=gray!20,
                                    minimum size=0.5cm]
	
\begin{tikzpicture}[>=latex,auto=left]
  
  \matrix[row sep=0.5cm,column sep=0.4cm] {
        \node (theta) [state] {$\theta_{i,t}$}; &
        &
        &
        \node (z1) [state] {$z_{i,j}^1$};       &
        \node (z2)   [state] {$z_{i,j}^2$};     &
        \node (zdots)  {$\cdots$}; &
        \node (zh) [state] {$z_{i,j}^h$};       &
        \node (d) [doc] {$d_{i,j}$}; &
        &
        \\
        &
        &
        &
        &
        &
        &
        &
        &
        &\\
        \node (alpha) [state] {$\mathbf{\alpha}$}; &
        &
        &
         &
         \node (hide) {};  &
        &
       \node (beta) [state] {$\mathbf{\beta}$};
        &
        \node (phi) [state] {$\mathbf{\phi}_t$} ;
        &
               & \\
        };
    
   \path[->]
   (theta)  edge (z1)
   (theta)  edge[bend right=25] (z2)
   (z1) edge (z2)
   (z2) edge (zdots)
   (zdots) edge (zh)
   (theta) edge[bend right=27] (zh)
   (zh) edge (d)
   (alpha) edge (theta)
   (phi) edge (d)
   (beta) edge (phi)
   ;
          
\node [
 	rectangle,thick, draw,
	left=-1.25cm of theta,
	minimum width=1.4cm,
	minimum height=1.4cm,
] (c1) {};    

\node[align=right, yshift = -0.5cm, xshift = -0.5cm] at (c1.east) {$T-\tau$};

\node [
 	rectangle,thick, draw, 
	right=0.2cm of c1,
	minimum width=6cm,
	minimum height=1.4cm
] (c2) {};  

\node[align=right, yshift = -0.5cm, xshift=0cm] at (d.east) {$l_i$};
\node[align=right, yshift = -0.7cm, xshift=0.4cm] at (d.east) 
{$D$};
\node [
 	rectangle,thick, draw, 
	left=-8cm of c1,
	minimum width=8.2cm,
	minimum height=1.8cm,
] (c3) {};  
    
\node [
 	rectangle,thick, draw, 
       below=0.8cm of d,
	minimum width=1.2cm,
	minimum height=1cm
] (c3) {};  

\node[align=right, yshift = -0.25cm, xshift=0.15cm] at (phi.east) {$\tau$};
\end{tikzpicture}
\caption{Latent Dirichlet Allocation with Topic Tree \hide{\xueqing{need modify and resize}}}
\label{fig:model}
\end{figure}
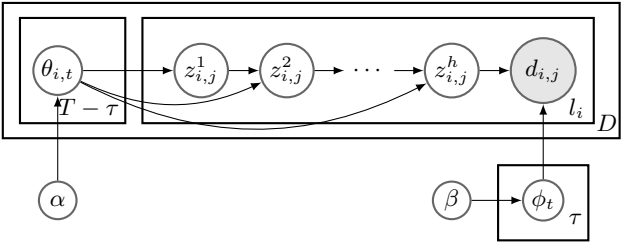

In our model, every document is modeled as a series of multinomial distributions: one multinomial distribution for every non-leaf topic over its child topics, representing the content bias towards the subtopics. For example, in Figure~\ref{fig:eg}, there are 4 non-leaf topics: $o,o/1,o/2$ and $o/3$. So a document $d_i$ is associated with 4 multinomial topic distributions: $\theta_{i,o}$ over its 3 children, and $\theta_{i,o/1},\theta_{i,o/2},\theta_{i,o/3}$ over their 2 children each. When the height of the hierarchy $H=1$, it reduces to the flat LDA model, because only the root is a non-leaf node. Each multinomial distribution $\theta_{i,t}$ is generated from a Dirichlet prior $\alpha_t$. $\alpha_{t,z}$ represents the corpus bias towards $z$-th child of topic $t$, and $\alpha_{t,0}=\sum_{z=1}^{C_t}\alpha_{t,z}$.

For every leaf topic node $t$, $C_t=0$, and a multinomial distribution $\phi_t$ over words is generated from another Dirchlet prior $\beta$. These word distributions are shared by the entire corpus.

To generate a word $d_{i,j}$, we first sample a path from the root to a leaf node $o/z_{i,j}^1/z_{i,j}^2/\cdots/z_{i,j}^h$. The nodes along the path are sampled one by one, starting from the root. Each time one child $z_{i,j}^{k}$ is selected from all children of $o/z_{i,j}^1/\cdots/z_{i,j}^{k-1}$, according to the multinomial $\theta_{i,o/z_{i,j}^1/\cdots/z_{i,j}^{k-1}}$. When a leaf node is reached, the word is generated from the multinomial distribution $\phi_{o/z_{i,j}^1/z_{i,j}^2/\cdots/z_{i,j}^h}$. Note that the length of the path $h$ is not necessary to be equal for all documents, if not all leaf nodes are on the same level.

The whole generative process is illustrated in Figure~\ref{fig:model}. Table~\ref{tb:notations} collects the notations. \hide{$\tau$ is the number of leaf nodes. }

\smallskip
\noindent
\textbf{A feature supporting recursive construction.} For every non-leaf topic node, we can derive a word distribution by marginalizing their child topic word distributions:
\begin{equation}
\label{eq:m1}
\phi_{t,x}=p(x|t)=\sum_{z=1}^{C_t}p(z|t)p(x|t/z)=\sum_{z=1}^{C_t}\frac{\alpha_{t,z}}{\alpha_{t,0}}\phi_{t/z,x}
\end{equation}
So in our model, the word distribution $\phi_t$ for an internal node in the topic hierarchy can be seen as a mixture of its child topic word distributions. The Dirichlet prior $\alpha_{t}$ determines the mixing weight.

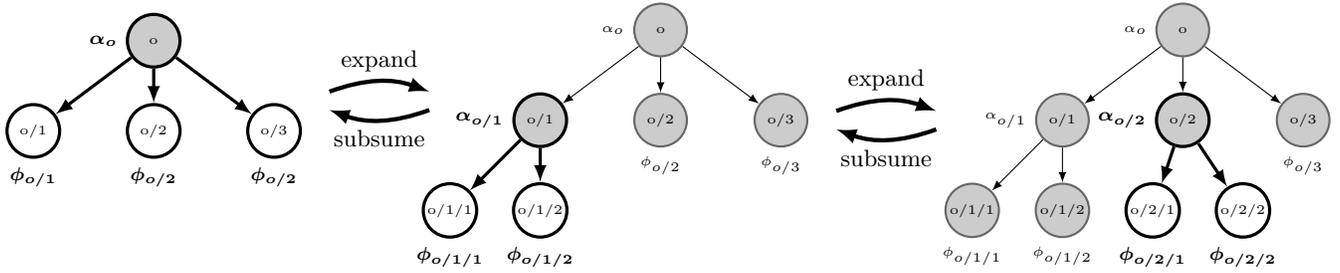
\begin{figure*}
\centering
\tikzstyle{new}=[circle,
                                    very thick,
                                    draw=black!100,
                                    inner sep = 1pt,
                                    minimum size=0.7cm]
                                    
\tikzstyle{constructing}=[circle,
                                    very thick,
                                    draw=black!100,
                                    fill=black!20,
                                    inner sep = 1pt,
                                    minimum size=0.7cm]
                                   
\tikzstyle{old}=[circle,
                                    thick,
                                    draw=black!60,
                                    fill=black!20,
                                    inner sep = 1pt,
                                    minimum size=0.7cm]
                                    
\tikzstyle{hide}=[circle,
                                    draw=none,
                                    inner sep = 1pt,
                                    minimum size=0.7cm]                                 
                                  
\begin{subfigure}{.29\linewidth}

 \begin{tikzpicture}[>=latex, scale=.4,auto=left, font = \tiny]

  \node (t0)[constructing][label=left:$\scriptsize\boldsymbol{\alpha_o}$] at (9,8) {o};
  \node (t1)[new][label=below:$\scriptsize\boldsymbol{\phi_{o/1}}$]  at (5,5)  {o/1};
  \node (t2)[new][label=below:$\scriptsize\boldsymbol{\phi_{o/2}}$] at (9, 5) {o/2};
  \node (t3)[new][label=below:$\scriptsize\boldsymbol{\phi_{o/2}}$] at (13, 5) {o/3};
  \node (t5)[hide][label=below:]  at (5,2) {};
  \node (t6)[hide][label=below:]  at (8,2) {};
  \node (t7)[hide][label=below:]  at (11,2) {};
  \node (t8)[hide] at (14,6) {};
  \node (t9)[hide] at (19,6) {};
   \path[->]
    (t0) edge[very thick] (t1)
    (t0) edge[very thick] (t2)
    (t0) edge[very thick] (t3)
    (t8) edge[ultra thick, bend left=20] node {\normalsize expand} (t9) 
    (t9) edge[ultra thick, bend left=20] node {\normalsize subsume} (t8) 
    ;
\end{tikzpicture}  
\end{subfigure}
\hfill
\begin{subfigure}{.38\linewidth}

 \begin{tikzpicture}[>=latex, scale=.4,auto=left, font=\tiny]

  \node (t0)[old][label=left:$\alpha_o$] at (9,8) {o};
  \node (t1)[constructing][label=left:$\scriptsize\boldsymbol{\alpha_{o/1}}$]  at (5,5)  {o/1};
  \node (t2)[old][label=below:$\phi_{o/2}$] at (9, 5) {o/2};
  \node (t3)[old][label=below:$\phi_{o/3}$] at (13, 5) {o/3};
  \node (t4)[new] at (2, 2) [label=below:\scriptsize$\boldsymbol{\phi_{o/1/1}}$]{o/1/1};
  \node (t5)[new] at (5, 2) [label=below:\scriptsize$\boldsymbol{\phi_{o/1/2}}$]{o/1/2};
 \node (t6)[new] at (8, 2) [hide][label=below:]{};
  \node (t7)[new] at (11, 2) [hide][label=below:]{};
  \node (t8)[hide] at (14,5) {};
  \node (t9)[hide] at (19,5) {};
  
   \path[->]
    (t0) edge (t1)
    (t0) edge (t2)
    (t0) edge (t3)
    (t1) edge[very thick]  (t4)
    (t1) edge[very thick]  (t5)
    (t8) edge[ultra thick, bend left=20] node {\normalsize expand} (t9) 
    (t9) edge[ultra thick, bend left=20] node {\normalsize subsume} (t8) 
    ;

\end{tikzpicture} 
\end{subfigure}
\hfill
\begin{subfigure}{.3\linewidth}
 \begin{tikzpicture}[>=latex, scale=.4,auto=left,font = \tiny]

  \node (t0)[old][label=left:$\alpha_o$] at (9,8) {o};
  \node (t1)[old][label=left:$\alpha_{o/1}$]  at (5,5)  {o/1};
  \node (t2)[constructing][label=left:$\scriptsize\boldsymbol{\alpha_{o/2}}$] at (9, 5) {o/2};
  \node (t3)[old][label=below:$\phi_{o/3}$] at (13, 5) {o/3};
  \node (t4)[old] at (2, 2) [label=below:$\phi_{o/1/1}$]{o/1/1};
  \node (t5)[old] at (5, 2) [label=below:$\phi_{o/1/2}$]{o/1/2};
  \node (t6)[new] at (8, 2) [label=below:\scriptsize$\boldsymbol{\phi_{o/2/1}}$]{o/2/1};
  \node (t7)[new] at (11, 2) [label=below:\scriptsize$\boldsymbol{\phi_{o/2/2}}$]{o/2/2};

    \path[->]
    (t0) edge (t1)
    (t0) edge (t2)
    (t0) edge (t3)
    (t1) edge (t4)
    (t1) edge (t5)
    (t2) edge[very thick]  (t6)
    (t2) edge[very thick]  (t7)
    ;

\end{tikzpicture}  
\end{subfigure}

 \caption{An illustration of recursive topical hierarchy construction. The construction order is from left to right. Each time one leaf topic node is expanded into several child topics (unshaded) and the relevant parameters (in bold) are estimated. The same figure explains \emph{subsumption} relationship: A tree on the left is subsumed by a tree on the right}
 \label{fig:recursion}
\end{figure*}

A topical hierarchy $\mathcal{T}$ is parameterized by $\alpha_t(\mathcal{T})$ where $C_t(\mathcal{T})>0$, and $\phi_t(\mathcal{T})$ where $C_t(\mathcal{T})=0$. We define a topical hierarchy $\mathcal{T}_1$ to be \emph{subsumed} by $\mathcal{T}_2$, 
if there is a mapping $\kappa$ from node $t$ in $\mathcal{T}_1$ to node $t'$ in $\mathcal{T}_2$, such that for every node $t$ in $\mathcal{T}_1$, $\pi_t(\mathcal{T}_1)=\pi_{\kappa(t)}(\mathcal{T}_2)$, and one of the following is true:
\begin{enumerate}
	\item $C_t(\mathcal{T}_1)=C_{\kappa(t)}(\mathcal{T}_2)>0$ and $\alpha_t(\mathcal{T}_1)=\alpha_{\kappa(t)}(\mathcal{T}_2)$; or
	\item $C_t(\mathcal{T}_1)=0$ and $\phi_{t}(\mathcal{T}_1)=\phi_{\kappa(t)}(\mathcal{T}_2)$.
\end{enumerate}
In words, a subsumed tree is obtained by removing all the descendants of some nodes in a larger tree, and absorbing the word distributions of the descendants into the new leaf nodes. 
In Figure~\ref{fig:recursion}, we show three trees and each tree is subsumed by the one on its right.
The subsumed tree retains equivalent high-level topic information of a larger tree, and can be recovered before we recover the larger tree. This idea allows us to recursively construct the whole hierarchy, which distinguishes our method from all-at-once construction methods. We present the recursive construction approach with justification in the next section.

\section{The STROD Framework}\label{sec:method}
We propose a Scalable Tensor Recursive Orthogonal Decomposition (STROD) framework for topical hierarchy construction, the first that meets all the criteria in Section~\ref{sec:property}.
It uses tensor (hypermatrix) decomposition to perform moment-based inference of the hierarchical topic model proposed in Section~\ref{sec:model} recursively.
\subsection{Moment-based Inference}\label{sec:moment}
The central idea of the inference method is based on the \emph{method of moments}, instead of \emph{maximum likelihood}. It enables tractable computations to estimate the parameters.

In statistics, the \emph{population moments} are expected values of powers of the random variable under consideration. The method of moments derives equations that relate the population moments to the model parameters. Then, it collects empirical population moments from observed samples, and solve the equations using the sample moments in place of the theoretical population moments. In our case, the random variable is the word occurring in each document. The population moments are expected occurrences and co-occurrences of the words.  They are related to the model parameters $\alpha$ and $\phi$. We can collect empirical population moments from the corpus, and estimate $\alpha$ and $\phi$ by fitting the empirical moments with theoretical moments. One particular computational advantage is that the inference only relies on the empirical population moments (word co-occurrence statistics). They compress important information from the full data, and require only one scan of the data to collect.

The idea is promising, but not straightforward to apply to our model. The challenge is that the same population moments can be expressed by parameters on different levels. For the example in Figure~\ref{fig:eg}, we can derive equations of the population moments (expected word co-occurrences) based on the model parameters associated with $t1,t2$ and $t3$, or based on those with $t4-t9$. Solving these equations independently will find 3 general topics and 6 more specific topics, but will neither reveal their relationship, nor guarantee the existence of the relationship.

Below we present a recursive inference method \hide{that handles this issue }step by step and justify its correctness.

\smallskip
\noindent
\textbf{1. Conditional population moments}.
We consider the population moments \emph{conditioned} on a non-leaf topic $t$. The first order moment is the expectation of word $x$'s occurrence given that it is drawn from topic $t$'s descendant. We have 
$p(x|t,\alpha) = \sum_{z=1}^{C_t}\frac{\alpha_{t,z}}{\alpha_{t,0}}\phi_{t/z,x}$ according to Equation~(\ref{eq:m1}).

The second order moment is the expectation of the co-occurrences of two words $x_1$ and $x_2$ given that they are both drawn from topic $t$'s descendants. \hide{Integrating over the document topic distribution $\theta_{i}$, we have:}
\begin{align}
\label{eq:m2}
\small
\begin{split}
	p(x_1,x_2|t,t,\alpha) 
	= &\sum_{z_1\neq z_2}\frac{\alpha_{t,z_1}\alpha_{t,z_2}}{\alpha_{t,0}(\alpha_{t,0}+1)}\phi_{t/z_1,x_1}\phi_{t/z_2,x_2}\\
	+ &\sum_{z=1}^{C_t}\frac{\alpha_{t,z}(\alpha_{t,z}+1)}{\alpha_{t,0}(\alpha_{t,0}+1)}\phi_{t/z,x_1}\phi_{t/z,x_2}
\end{split}
\end{align}

Likewise, we can derive the third order moment as the expectation of co-occurrences of three words $x_1,x_2$ and $x_3$ given that they are all drawn from topic $t$'s descendants:
\begin{align}
\label{eq:m3}
\small
	\begin{split}
		&p(x_1,x_2,x_3|t,t,t,\alpha) \\
		= & \sum_{z_1\neq z_2\neq z_3 \neq z_1}\frac{\alpha_{t,z_1}\alpha_{t,z_2}\alpha_{t,z_3}}{\alpha_{t,0}(\alpha_{t,0}+1)(\alpha_{t,0}+2)}\phi_{t/z_1,x_1}\phi_{t/z_2,x_2}\phi_{t/z_3,x_3}\\
		+ & \sum_{z_1\neq z_2}\frac{\alpha_{t,z_1}\alpha_{t,z_2}(\alpha_{t,z_1}+1)}{\alpha_{t,0}(\alpha_{t,0}+1)(\alpha_{t,0}+2)}(\phi_{t/z_1,x_1}\phi_{t/z_1,x_2}\phi_{t/z_2,x_3}\\
		+ & \phi_{t/z_1,x_1}\phi_{t/z_1,x_3}\phi_{t/z_2,x_2} + \phi_{t/z_1,x_3}\phi_{z_1,x_2}\phi_{z_2,x_1})\\
		+ & \sum_{z=1}^{C_t}\frac{\alpha_{t,z}(\alpha_{t,z}+1)(\alpha_{t,z}+2)}{\alpha_{t,0}(\alpha_{t,0}+1)(\alpha_{t,0}+2)}\phi_{t/z,x_1}\phi_{t/z,x_2}\phi_{t/z,x_3}
	\end{split}
\end{align}

These equations exhibit good opportunities for a recursive solution, because the moments conditioned on a topic $t$ can be expressed by only the Dirichlet prior and word distributions associated with its child topics. If these low order moments can uniquely determine the model parameters, we can use them to recover the child topics of every topic robustly, and by recursion, we can then construct the whole tree (Figure~\ref{fig:recursion}).

Fortunately, there is indeed a robust technique to recover the parameters from low order moments.

\smallskip
\noindent
\textbf{2. Tensor orthogonal decomposition}.
Anandkumar \etal~\cite{anandkumar2012} proved that with some mild non-degeneracy conditions, the following equations can be uniquely solved by a tensor orthogonal decomposition method:
\begin{align}
\small
	\label{eq:decomp}
	\begin{split}
	M_2 &= \sum_{z=1}^k \lambda_z v_z \otimes v_z,
	M_3 = \sum_{z=1}^k \lambda_z v_z \otimes v_z \otimes v_z		
	\end{split}
\end{align}
where $M_2$ is a $V\times V$ tensor (hence, a matrix) and $M_3$ is a $V\times V\times V$ tensor, $\lambda_z$ is an unknown positive value about the weight of $z$-th component $v_z$, which is an unknown $V$-dimensional vector. In words, both $M_2$ and $M_3$ can be decomposed into the same number of components, and each component is determined by a single vector.
The operator $\otimes$ denotes an outer product between tensors: if $A\in \real^{s_1\times\cdots\times s_p}$, and $B\in\real^{s_{p+1}\times\cdots\times s_{p+q}}$, then $A\otimes B$ is a tensor in $\real^{s_1\times\cdots\times s_{p+q}}$, and $[A\otimes B]_{i_1\dots i_{p+q}}=A_{i_1\dots i_p}B_{i_{p+1}\dots i_{p+q}}$.

To write Equations~(\ref{eq:m1})-(\ref{eq:m3}) in this form, we define:
\nop{
\begin{equation}
\label{eq:M1}
M_1(t) = \sum_{z=1}^{C_t}\frac{\alpha_{t,z}}{\alpha_{t,0}}\phi_{t/z}
\end{equation}
\begin{equation}
\label{eq:E2}
	E_2(t) = [p(x_1,x_2|t,t,\alpha)]_{V\times V} 
\end{equation}
\begin{equation}
\label{eq:M2}
	M_2(t) = (\alpha_{t,0}+1) E_2(t) - \alpha_{t,0} M_1(t)\otimes M_1(t) \\
	E_3(t) = [p(x_1,x_2,x_3|t,t,t,\alpha)]_{V\times V\times V} 
\end{equation}
}
\small
\begin{align}
\label{eq:M1}
	M_1(t) & = \sum_{z=1}^{C_t}\frac{\alpha_{t,z}}{\alpha_{t,0}}\phi_{t/z}\\
	E_2(t) & = [p(x_1,x_2|t,t,\alpha)]_{V\times V} \\
\label{eq:M2}
	M_2(t) & = (\alpha_{t,0}+1) E_2(t) - \alpha_{t,0} M_1(t)\otimes M_1(t) \\
	E_3(t) & = [p(x_1,x_2,x_3|t,t,t,\alpha)]_{V\times V\times V} \\
	\begin{split}
	U_1(t) & = E_2(t)\otimes M_1(t), \\
  & U_2(t) = \Omega(U_1(t),1,3,2), U_3(t)=\Omega(U_1(t),2,3,1)
	\end{split}\\
	\begin{split}
		M_3(t) & = \frac{(\alpha_{t,0}+1)(\alpha_{t,0}+2)}{2}E_3(t)+ \alpha_{t,0}^2 M_1\otimes M_1\otimes M_1\\
		& - \frac{\alpha_{t,0}(\alpha_{t,0}+1)}{2}[U_1(t)+U_2(t)+U_3(t)]	
	\end{split}
\label{eq:M3}
\end{align}
\normalsize
where $\Omega(A,a,b,c)$ permutes the modes of tensor $A$, such that $\Omega(A,a,b,c)_{i_1,i_2,i_3}=A_{i_a,i_b,i_c}$.
It follows that:
\small
	$$M_2(t) = \sum_{z=1}^{C_t}\frac{\alpha_{t,z}}{\alpha_{t,0}}\phi_{t/z}\otimes\phi_{t/z},
          M_3(t) = \sum_{z=1}^{C_t}\frac{\alpha_{t,z}}{\alpha_{t,0}}\phi_{t/z}\otimes\phi_{t/z}\otimes\phi_{t/z}$$
\normalsize
So they fit Equation~(\ref{eq:decomp}) nicely, and intuitively. If we decompose $M_2(t)$ and $M_3(t)$, the $z$-th component is determined by the child topic word distribution $\phi_{t/z}$, and its weight is $\frac{\alpha_{t,z}}{\alpha_{t,0}}$, which is equal to $p(t/z|t,\alpha_t)$.

\begin{algorithm}
\caption{Tensor Orthogonal Decomposition (TOD)} 
\label{alg:tod}
\small
\SetLine
\KwIn{Tensor $M_2\in \real^{V\times V}$, $M_3\in \real^{V\times V\times V}$, number of components $k$, number of outer and inner iterations $N$ and $n$}
\KwOut{The decomposed components $(\lambda_z,v_z), z=1,\dots,k$}
\BlankLine
Compute $k$ orthonormal eigenpairs $(\sigma_z,\mu_z)$ of $M_2$\;\nllabel{line:eigen}

Compute a whitening matrix $W=M\Sigma^{-\frac{1}{2}}$\; \tcp{$M=[\mu_1,\dots,\mu_k],\Sigma=diag(\sigma_1,\dots,\sigma_k),W^TM_2W=I$} \nllabel{line:whiten}

Compute $(W^T)^{+}=M\Sigma^{\frac{1}{2}}$ \; \tcp{the Moore-Penrose pseudoinverse of $W^T$} \nllabel{line:inverse}

Compute a $k\times k \times k$ tensor $\widetilde{T}=M_3(W,W,W)$\;  \tcp{$\widetilde{T}_{i1,j1,k1}=\sum_{i2,j2,k2}(M_3)_{i2,j2,k2}W_{i2,i1}W_{j2,j1}W_{k2,k1}$}
\nllabel{line:project}

\For{$z=1..k$}
{
\nllabel{line:powerbegin}
	$\lambda^* \gets 0$ \tcp*{the largest eigenvalue so far}

  \For{$outIter=1..N$}
  {
  	$v \gets $ a random unit-form vector\;
	\lFor{$innerIter=1..n$}
	{
		$v \gets \frac{\widetilde{T}(I,v,v)}{||\widetilde{T}(I,v,v)||}$
	}
		\tcp{power iteration update}\nllabel{line:power}
	\lIf{$\widetilde{T}(v,v,v)>\lambda^*$}
		{($\lambda^*,v^*)\gets (\widetilde{T}(v,v,v)$,v)}
		\tcp{choose the largest eigenvalue}
		\nllabel{line:pick}
  }
   $\lambda_z = \frac{1}{(\lambda^*)^2}$, $v_z= \lambda_z(W^T)^+ v^*$\; \tcp{recover eigenpair of the original tensor} \nllabel{line:recover}
   $\widetilde{T}\gets \widetilde{T}-\lambda^* v^*\otimes v^*\otimes v^*$ \tcp*{deflation} \nllabel{line:deflate}
} \nllabel{line:powerend}

	\Return $(\lambda_z,v_z),z=1,\dots,k$

\normalsize
\end{algorithm}

Algorithm~\ref{alg:tod} outlines the tensor orthogonal decomposition method for recovering the components. It can be partitioned into two parts:

\begin{enumerate}
\item Lines~\ref{line:eigen} to \ref{line:project} project the large tensor $M_3\in \real^{V\times V\times V}$ into a smaller tensor $\widetilde{T}\in\real^{k\times k\times k}$. $\widetilde{T}$ is not only of smaller size, but can be decomposed into an orthogonal form: $\widetilde{T}=\sum_{z=1}^k \widetilde{\lambda_i}\widetilde{v_i}^{\otimes 3}$. $\widetilde{v_i},i=1,\dots,k$ are orthonormal vectors in $\real^{k}$. This is assured by the whitening matrix $W$ calculated in Line~\ref{line:whiten}, which satisfies $W^TM_2W=I$.
\item Lines~\ref{line:powerbegin} to \ref{line:powerend} perform orthogonal decomposition of $\widetilde{T}$ via a power iteration method. The orthonormal eigenpairs $(\widetilde{\lambda_z},\widetilde{v_z})$ are found one by one. To find one such pair, the algorithm randomly starts with a unit-form vector $v$, runs power iteration (Line~\ref{line:power}) for $n$ times, and records the candidate eigenpair.  This process further repeats by $N$ times, starting from different unit-form vectors, and the candidate eigenpair with the largest eigenvalue is picked (Line~\ref{line:pick}). After an eigenpair is found, the tensor $\widetilde{T}$ is deflated by the found component (Line~\ref{line:deflate}), and the same power iteration is applied to it to find the next eigenpair. After all the $k$ orthonormal eigenpairs $(\widetilde{\lambda_z},\widetilde{v_z})$ are found, they can be used to uniquely determine the $k$ target components $(\lambda_z,v_z)$ (Line~\ref{line:recover}).
\end{enumerate}

The following theorem ensures that the decomposition is unique and fast.
\begin{theorem}
\label{the:unique}
Assume $M_2$ and $M_3$ are defined as in Equation~\ref{eq:decomp}, $\lambda_z>0$, and the vectors $v_z$'s are linearly independent and have unit-form, then Algorithm~\ref{alg:tod} returns exactly the same set of $(\lambda_z, v_z)$. Furthermore, the power iteration step of Line~\ref{line:power} converges in a quadratic rate.
\end{theorem}
This theorem is essentially a combination of Theorem 4.3 and Lemma 5.1 in Anandkumar \etal~\cite{anandkumar2012}. See Appendix A for more discussion about it. The importance of Theorem~\ref{the:unique} is that it allows us to use moments only up to the third order to recover the exact components, and the convergence is fast.

\smallskip
\noindent
\textbf{3. Recursive decomposition}. With Algorithm~\ref{alg:tod} as a building block, we can divide and conquer the inference of the whole model. We devise Algorithm~\ref{alg:rtod}, which recursively infers model parameters in a top-down manner. Taking any topic node $t$ as input, it computes the conditional moments $M_2(t)$ and $M_3(t)$. If $t$ is not root, they are computed from the parent topic's moments and estimated model parameters. For example, according to Bayes's theorem,
\begin{align}
\small
\begin{split}
	& [E_2(t)]_{x_1,x_2} = p(x_1,x_2|t,t,\alpha) \propto p(x_1,x_2,t,t|\pi_t,\pi_t,\alpha)\\
	 = & ~p(x_1,x_2|\pi_t,\pi_t,\alpha)p(t,t|x_1,x_2,\pi_t,\pi_t,\alpha) \\	
	= & ~[E_2(\pi_t)]_{x_1,x_2}{\alpha_{\pi_t,\chi_t}(\alpha_{\chi_t,z}+1)\phi_{t,x_1}\phi_{t,x_2}}\\
	/& \left(\sum_{z=1}^{C_{\pi_t}}\alpha_{\pi_t,z}(\alpha_{\pi_t,z}+1)\phi_{\pi_t/z,x_1}\phi_{\pi_t/z,x_2}\right.\\
	& +\left.\sum_{z_1\neq z_2}\alpha_{\pi_t,z_1}\alpha_{\pi_t,z_2}\phi_{\pi_t/z_1,x_1}\phi_{\pi_t/z_2,x_2} \right)
\end{split}
\end{align}
\normalsize
Other quantities in Equations~(\ref{eq:M1})-(\ref{eq:M3}) can be computed similarly.
Then it performs tensor decomposition and recovers the parameter $\alpha_t$ and $\phi_{t/z}$ for each child topic. It then enumerates its children and makes recursive calls with each of them as input. The recursion stops when reaching leaf nodes, where $C_t=0$.  A call of Algorithm~\ref{alg:rtod} with the root $o$ as input will construct the entire hierarchy.

\begin{algorithm}
\caption{Recursive Tensor Orthogonal Decomposition (RTOD)} 
\label{alg:rtod}
\small
\SetLine
\KwIn{topic $t$, number of outer and inner iterations $N,n$}
\BlankLine
Compute $M_2(t)$ and $M_3(t)$\; \tcp{only relies on $t$'s ancestors}\nllabel{line:moment}

$(\lambda_z,v_z) \gets TOD(M_2(t),M_3(t),C_t,N,n)$\; 

$\alpha_{t,z}=\alpha_{t,0}\lambda_z, \phi_{t/z}=v_z$\;

\For{$z=1..C_t$}
{
	RTOD($t/z,N,n$) \;
	\tcp{Recursion, get the parameters for the subtree rooted at each child}
} 
\normalsize
\end{algorithm}

The correctness of this recursive algorithm is permitted by the special structure of our model. In particular, we have the following theorem.
\begin{theorem}
The RTOD algorithm (Algorithm~\ref{alg:rtod}) has both robust recovery and robust revision property.
\end{theorem}
The robust recovery property follows Theorem~\ref{the:unique}, plus the fact that the conditional moments of a topic can be expressed by only the Dirichlet prior and word distributions associated with its child topics. The robust revision property is due to conditional independence during the recursive construction procedure: i) once a topic $t$ has been visited in the algorithm, the construction of its children is independent of each other; and ii) the conditional moments $M_2(t)$ and $M_3(t)$ can be computed independently of $t$'s descendants. In fact, it leads to a stronger claim.
\begin{corollary}
If $\mathcal{T}_1$ is subsumed by $\mathcal{T}_2$ with the mapping $\kappa(\cdot)$, then the RTOD algorithm on $\mathcal{T}_1$ and $\mathcal{T}_2$ returns identical parameters for $\mathcal{T}_1$ and $\kappa(\mathcal{T}_1)$.
\end{corollary}

Therefore, the tree topology can be expanded or varied locally with minimal revision to the inferred topics. This is in particular useful when the structure of the topic tree is not fully determined in the beginning. The recursive construction offers users a chance to see the construction results and interact with the topic tree expansion or its local variations by deciding on the number of topics. \hide{We discuss how to automatically select the number of topics in Appendix B\hide{Section~\ref{sec:hyper}}.}

\subsection{Scalability Improvement}\label{sec:scale}
Although Algorithms~\ref{alg:tod} and \ref{alg:rtod} are robust, they are not scalable.  The orthogonal decomposition of the tensor $\widetilde{T}\in\real^{k\times k\times k}$ (Lines~\ref{line:powerbegin}-\ref{line:powerend}) is efficient, because $k$ is small. However, the bottleneck of the computation is preparing the tensor $\widetilde{T}$, including Line~\ref{line:moment} and Lines~\ref{line:eigen} to \ref{line:project}. They involve the creation of a dense tensor $M_3\in\real^{V\times V\times V}$, and the time-consuming operation $M_3(W,W,W)$. Since $V$ is usually tens of thousands or larger, it is impossible to store such a tensor in memory and perform the tensor product operation. In fact, even the second order moment $M_2\in \real^{V\times V}$ is dense and large, challenging both space and time efficiency already.

Anandkumar~\cite{anandkumar2012} discusses a plausible way to solve the space challenge, by avoiding explicit creation of the tensors $M_3$ and $\widetilde{T}$. It suggests going through the document-word occurrence data for computing the power iterations. This requires as many times of data passes as other inference methods, if not more. Therefore, its scalability will still be unsatisfactory.

We make key contributions to solving the challenge in a different approach.  We avoid explicit creation of both tensor $M_3$ and $M_2$.  
But we do explicitly create $\widetilde{T}$ since it is memory efficient.  Therefore, the efficient power iteration updates remain as in Algorithm~\ref{alg:tod}. 
Utilizing the special structure of the tensors in our problem, we show that $\widetilde{T}$ can be created by passing the data only twice, without incurring creations of any dense $V^2$ or $V^3$ tensors.

\smallskip
\noindent
\textbf{1. Avoid creating $\mathbf{M_2}$}. For ease of discussion, we omit the conditional topic $t$ in the notation of this and next subsection. According to Equation~(\ref{eq:M2}), $M_2=(\alpha_{0}+1) E_2 - \alpha_{0} M_1\otimes M_1 $. $E_2$ is a sparse symmetric matrix because only two words co-occurring in one document will contribute to the empirical estimation of $E_2$. However, $M_1\otimes M_1$ is a full $V$ by $V$ matrix. We would like to compute the whitening matrix $W$ without explicit creation of $M_2$.

First, we notice that $M_1$ is in the \emph{column space} of $M_2$ (\ie, $M_1$ is a linear combination of $M_2$'s column vectors), so $E_2$ has the same column space $S$ as $M_2$. Also, since $M_2=\sum_{z=1}^k \lambda_z v_z \otimes v_z$ is positive definite, so is $E_2=\frac{1}{\alpha_0+1}(M_2+\alpha_0M_1\otimes M_1)$.
Let $E_2=U\Sigma_1U^T$ be its spectral decomposition, where $U\in\real^{V\times k}$ is the matrix of $k$ eigenvectors, and $\Sigma_1\in\real^{k\times k}$ is the diagonal eigenvalue matrix. 
The $k$ column vectors of $U$ form an orthonormal basis of $S$. $M_1$'s representation in this basis is $M'_1=U^TM_1$. Now, $M_2$ can be written as:
	$$M_2=U[(\alpha_0+1)\Sigma_1-\alpha_0 M'_1\otimes M'_1]U^T$$
So, a second spectral decomposition can be performed on $M'_2=(\alpha_0+1)\Sigma_1-\alpha_0 M'_1\otimes M'_1$, as $M'_2=U'\Sigma U'^T$. Then we have:
	$$M_2=UU'\Sigma(UU')^T$$
Therefore, we effectively obtain the spectral decomposition of $M_2=M\Sigma M^T$ without creating $M_2$. Not only the space requirement is reduced (from a dense $V\times V$ matrix to a sparse matrix $E_2$), but also the time for spectral decomposition. If we perform spectral decomposition for $M_2$ directly, it requires $O(V^3)$ time complexity. However, using the twice spectral decomposition trick above, we just need to compute the first largest $k$ eigenpairs for a sparse matrix $E_2$, and a spectral decomposition for a small matrix $M'_2\in\real^{k\times k}$.  The first decomposition can be done efficiently by a power iteration method or other more advanced algorithms~\cite{tang2013}.  The time complexity is roughly $O(k\|E_2\|_0)$, where $\|E_2\|_0$ is the number of non-zero elements in $E_2$.  The second decomposition requires $O(k^3)$ time, which can be regarded as constant since $k<=K\approx 10$.

\smallskip
\noindent
\textbf{2. Avoid creating $\mathbf{M_3}$}. 
The idea is to directly compute $\widetilde{T}=M_3(W,W,W)$ without creating $M_3$.  This is possible due to the distributive law: $(A+B)(W,W,W)=A(W,W,W)+B(W,W,W)$. The key is to decouple $M_3$ as a summation of many different tensors, such that the computation of the product between each tensor and $W$ is easy.

We begin with the empirical estimation of $E_3$. Suppose we use $c_{i,x}$ to represent the count of word $x$ in document $d_i$. Then $E_3$ can be estimated by averaging all the 3-word triples in each document:
\begin{align}
\small
	\begin{split}
	E_3 & = \frac{1}{D}\left[A_1-A_2-\Omega(A_2,2,1,3)-\Omega(A_2,2,3,1)+2A_3\right]\\
	A_1 & = \sum_{i=1}^{D}\frac{1}{l_i(l_i-1)(l_i-2)}c_i\otimes c_i\otimes c_i\\
	A_2 & = \sum_{i=1}^D\frac{1}{l_i(l_i-1)(l_i-2)}c_i\otimes diag(c_i)\\
	A_3 & = \sum_{i=1}^D\frac{1}{l_i(l_i-1)(l_i-2)}tridiag(c_i)
	\end{split}
\end{align}
where $tridiag(v)$ is a tensor with vector $v$ on its diagonal: $tridiag(v)_{i,i,i}=v_i$.
Let $s_i=\frac{1}{l_i(l_i-1)(l_i-2)}$.
From the fact $(v\otimes v\otimes v)(W,W,W)=(W^Tv)\otimes(W^Tv)\otimes(W^Tv)=(W^Tv)^{\otimes 3}$, we can derive:
\small
\begin{align}
\label{eq:A1}
	A_1(W,W,W)&=\sum_{i=1}^D s_i(W^T c_i)^{\otimes 3}
\end{align}
\normalsize
Based on another fact, \small $(v\otimes M)(W,W,W)=(W^Tv)\otimes M(W,W) =(W^Tv)\otimes W^TMW$\normalsize, we can derive:\small
\begin{align}
\label{eq:A2}
	A_2(W,W,W) = \sum_{i=1}^D s_i (W^T c_i)\otimes W^T diag(c_i) W
\end{align}
\normalsize
Let $W^T_x$ be the $x$-th column of $W^T$, we have:\small
\begin{align}
\label{eq:A3}
	A_3(W,W,W)&=\sum_{x=1}^V\sum_{i=1}^D s_i c_{i,x}(W^T_x)^{\otimes 3}
\end{align}
\normalsize

So we do not need to explicitly create $E_3$ to compute $E_3(W,W,W)$. 
The time complexity using Equations~(\ref{eq:A1})-(\ref{eq:A3}) is $O(Lk^2)$, which is equivalent to $O(L)$ because $k$ is small.

Using the same trick, we can obtain:
\small
\begin{align}
\label{eq:U1}
	U_1(W,W,W) &= W^T E_2 W\otimes W^TM_1 \\
	(M1\otimes M1\otimes M1)(W,W,W) &= (W^T M_1)^{\otimes 3}
\end{align}
\normalsize
Equation~(\ref{eq:U1}) requires $O(k^2\|E_2\|_0 )$ time to compute, while $\|E_2\|_0$ can be large. We can further speed it up.

We notice that $W^T M_2 W=I$ by definition. Substituting $M_2$ with Equation~(\ref{eq:M2}), we have:
\small
\begin{align}
	W^T [(\alpha_{0}+1) E_2 - \alpha_{0} M_1\otimes M_1] W &= I
\end{align}
\normalsize
which is followed by:
\small
\begin{align}
\label{eq:fastu1}
	W^T E_2 W = \frac{1}{(\alpha_0+1)}[I+\alpha_0 (W^T M_1)^{\otimes 2}]
\end{align}
\normalsize
Pluging Equation~(\ref{eq:fastu1}) into (\ref{eq:U1}) further reduces the complexity of computing $U_1(W,W,W)$ to $O(Vk+k^3)$. $U_2(W,W,W)$ and $U_3(W,W,W)$ can be obtained by permuting $U_1(W,W,W)$'s modes, in $O(k^3)$ time.

Putting these together, we have the following fast computation of $\widetilde{T}=M_3(W,W,W)$ by passing the data once:
\begin{align}
\small
\label{eq:T}
	\begin{split}
		& \widetilde{T}=M_3(W,W,W)=	\frac{(\alpha_{t,0}+1)(\alpha_{t,0}+2)}{2}E_3(W,W,W)\\
		- & \frac{\alpha_{t,0}(\alpha_{t,0}+1)}{2}[(U_1+U_2+U_3)(W,W,W)]
		 + \alpha_{t,0}^2 (W^T M_1)^{\otimes 3}
	\end{split}
\end{align}
which requires $O(Lk^2+Vk^2+k^3)=O(L)$ time in total.

\smallskip
\noindent
\textbf{3. Estimation of empirical conditional moments.} To estimate the empirical conditional moments for topic $t$, we estimate the `topical' count of word $x$ in document $d_i$ as:
\begin{align}
\small
\label{eq:tc}
\begin{split}
	c_{i,x}(t)&=c_{i,x}p(t|x)
	=c_{i,x}(\pi_t)\frac{\alpha_{\pi_t,\chi_{t}}\phi_{t,x}}
	{\sum_{z=1}^{C_{\pi_t}}\alpha_{\pi_t,z}\phi_{\pi_t/z,x}}
\end{split}	
\end{align}
and $c_{i,x}(o)=c_{i,x}$. 
Then we can estimate $M_1$ and $E_2$ using these empirical counts:
\begin{align}
\small
\label{eq:E2}
\begin{split}
	M_1(t)&=\sum_{i=1}^D \frac{1}{l_i(t)} c_{i}(t) \\
	E_2(t)&=\sum_{i=1}^D \frac{1}{l_i(t)(l_i(t)-1)}	[c_i(t)\otimes c_i(t)-diag(c_i(t))]
\end{split}
\end{align}
where $l_i(t)=\sum_{x=1}^V c_{i,x}(t)$.  These enable fast estimation of empirical moments by passing data once.

Finally, we have a scalable tensor recursive orthogonal decomposition algorithm as outlined in Algorithm~\ref{alg:strod}.

\begin{algorithm}
\caption{Scalable Tensor Recursive Orthogonal Decomposition (STROD)} 
\label{alg:strod}
\small
\SetLine
\KwIn{topic $t$, number of outer and inner iterations $N,n$}
\BlankLine
Compute $M_1(t)$ and $E_2(t)$ according to Equation~(\ref{eq:E2})\;

Find $k$ largest orthonormal eigenpairs $(\sigma_z,\mu_z)$ of $E_2$\; \nllabel{line:spectral}

$M'_1=UM_1(t)$ \tcp*{$U=[\mu_1,\dots,\mu_k],\Sigma_1=diag(\sigma_1,\dots,\sigma_k)$}

Compute spectral decomposition for $M'_2=(\alpha_{t,0}+1)\Sigma_1-\alpha_{t,0} M'_1\otimes M'_1=U'\Sigma U'^T$\;

$M=UU',W=M\Sigma^{-\frac{1}{2}},(W^T)^{+}=M\Sigma^{\frac{1}{2}}$\;

Compute $\widetilde{T}=M_3(W,W,W)$ according to Equation~(\ref{eq:T})\;

Perform power iteration Line~\ref{line:powerbegin} to \ref{line:powerend} in Algorithm~\ref{alg:tod}\;

$\alpha_{t,z}=\alpha_{t,0}\lambda_z, \phi_{t/z}=v_z$\;

\For{$z=1..C_t$}
{
	STROD($t/z,N,n$) \;
} 
\normalsize
\end{algorithm}

We notice that the hyperparameter $\alpha_{t,0}$ and the number of child topics $C_t$ are needed to run the STROD algorithm.  
We discuss how to learn them automatically in Appendix B. 

\subsection{Phrase Mining and Ranking} \label{sec:phrase}
After the word distribution in each topic is inferred, we can then mine and rank topical phrases within each topic. The phrase mining and ranking in STROD largely follow CATHY~\cite{Wang13CATHY}. Here we briefly present the process.  

\smallskip
\noindent
\textbf{1. Mining.} In this work, a phrase is defined as a consecutive sequence of words. When representing a topic, only frequent phrases are of interest. \hide{CATHY uses a frequent pattern mining approach to mine phrases from short, content-representative documents.  In this work, we do not assume the documents are short. }So we treat each sentence as a `transaction', and each word as an `item', and mine frequent consecutive patterns as phrases of arbitrary lengths.  To filter out incomplete phrases (\eg, `vector machine' instead of `support vector machine') and frequently co-occurred words that do not make up a meaningful phrase (\eg, `learning classification'), CATHY defines two criteria \emph{completeness} and \emph{phraseness} to measure them.  Following that principle, we use a statistical test to decide 
quality phrases, and record the count $c_{i,P}$ of each phrase $P$ in each document $d_i$.

\smallskip
\noindent
\textbf{2. Ranking.} 
After the set of frequent phrases of mixed lengths is mined, they are ranked with regard to the representativeness of each topic in the hierarchy, based on two factors: \emph{popularity} and \emph{discriminativeness}. A phrase is \emph{popular} for a topic if it appears frequently in documents containing that topic (\eg, `information retrieval' has better popularity than `cross-language information retrieval' in the Information Retrieval topic).  A phrase is \emph{discriminative} of a topic if it is frequent only in the documents about that topic but not in those about other topics (\eg, `query processing' is more discriminative than `query' in the Databases topic).

We use the topic word distributions inferred from our model to estimate the `topical' count $c_{i,P}(t)$ of each phrase $P$ in each document $d_i$, in a similar way as we estimate the topical count of words in Equation~(\ref{eq:tc}):
\begin{align}
\label{eq:tcp}
\small
\begin{split}
	c_{i,P}(t)&=c_{i,P}(\pi_t)p(t|P,\pi_t)
	=c_{i,P}(\pi_t)\frac{\alpha_{\pi_t,\chi_{t}}\prod_{x\in P}\phi_{t,x}}
	{\sum_{z=1}^{C_{\pi_t}}\alpha_{\pi_t,z}\prod_{x\in P}\phi_{\pi_t/z,x}}
\end{split}
\end{align}

Let the conditional probability $p(P|t)$ be the probability of ``randomly choose a document and a phrase that is about topic $t$, the phrase is $P$\hide{seeing phrase $P$ in a document given that it is about topic $t$}.'' It can be estimated as $p(P|t)=\frac{1}{D}\sum_{i=1}^{D}\frac{c_{i,P}(t)}{\sum_{P'}c_{i,P'}(t)}$.
The popularity of a phrase in a topic $t$ can be quantified by $p(P|t)$. The discriminativeness can be measured by the log ratio of the probability $p(P|t)$ conditioned on topic $t$,
and 
the probability $p(P|\pi_t)$ conditioned on its parent topic $\pi_t$:
$\log\frac{p(P|t)}{p(P|\pi_t)}$.

As studied in Wang \etal~\cite{Wang13CATHY}, a good way to combine these two factors is to use their product:
\begin{align}
  r_t(P) = p(P|t)\log\frac{p(P|t)}{p(P|\pi_t)}
\end{align}
which has an information-theoretic meaning: the pointwise KL-divergence between two probabilities.  Finally, we use $r_t(P)$ to rank phrases in topic $t$ in the descending order.

\section{Experiments}\label{sec:exp}
In this section we first introduce the datasets and the methods used for comparison, and then describe our evaluation on \emph{scalability}, \emph{robustness}, and \emph{interpretability}.


\smallskip \noindent
\textbf{Datasets}. Our performance study is on four datasets:

\vspace*{-1.0ex}
\begin{itemize}
\parskip -0.2ex
\item {DBLP title}:  A set of titles of recently published papers in DBLP\footnote{\small \url{http://www.dblp.org}}. The set has 1.9M titles, 152K unique words, and 11M tokens. \hide{We minimally pre-processed the dataset by removing all stopwords from the titles, resulting in a collection of 33,313 titles consisting of 18,598 unique terms. We also use a subset DDMIN in the areas related to Databases, Data Mining, Machine Learning, Information Retrieval, and Natural Language Processing, as used in~\cite{Wang13CATHY}, for the sake of quality evaluation. }
\item {CS abstract}:  A dataset of computer science paper abstracts from Arnetminer\footnote{\small \url{http://www.arnetminer.org}}. The set has 529K papers, 186K unique words, and 39M tokens.
\item {TREC AP news}: A TREC news dataset (1998). It contains 106K full articles, 170K unique words, and 19M tokens.
\item {Pubmed abstract}: A dataset of life sciences and biomedical topic. We crawled 1.5M abstracts\footnote{\small \url{http://www.ncbi.nlm.nih.gov/pubmed}} from Jan. 2012 to Sep. 2013. The dataset has 98K unique words after stemming and 169M tokens.

\end{itemize}
We remove English stopwords from all the documents. Documents shorter than 3 tokens are not used for computing the moments because we rely on up to third order moments.

\smallskip
\noindent
\textbf{Methods for Comparison.}\label{subsec:methods_compare}
%
%
Our method is compared with the following topical hierarchy construction methods.

\vspace*{-1.0ex}
\begin{itemize}
\parskip -0.4ex
\item {hPAM -- parametric hierarchical topic model}. The hierarchical Pachinko Allocation Model~\cite{Mimno07} is a state-of-the-art parametric hierarchical topic modeling approach. hPAM outputs a specified number of supertopics and subtopics, as well as the associations between them.
\item {nCRP -- nonparametric hierarchical topic model}. \hide{As a second baseline, we use a nonparametric hierarchical topic modeling with nested Chinese Restaurant Process~\cite{Griffiths04}. }Although more recently published nonparametric models have more capability in document modeling, their scalability is worse than nested Chinese Restaurant Process~\cite{Griffiths04}\hide{\footnote{Paisley \etal~\cite{paisley2012} proposed a stochastic inference algorithm that is more scalable, but it is not published yet}}. So we choose nCRP to represent this category. It outputs a tree with a specified height, but the number of topics is determined by the algorithm.  A hyperparameter can be tuned to generate more or fewer topics. In our experiment we tune it to generate an approximately identical number of topics as other methods.
\item {splitLDA -- recursively applying LDA}. We implement a recursive method described by Pujara and Skomoroch~\cite{pujara:nips12}. They use LDA to infer topics for each level, and split the corpus according to the inferred results to produce a smaller corpus for inference with the next level. \hide{ The split of corpus is heuristic and has no model-wise justification, but }This heuristic method has the best known scalability so far\hide{, especially when combined with parallel implementation of LDA inference algorithms}. For fair comparison of the fundamental computational complexity of different algorithms, we do not use any parallel implementation for all the methods. So we implement splitLDA on top of a fast single-machine LDA inference algorithm~\cite{yao2009efficient}.
\item {CATHY -- recursively clustering word co-occurrence networks}. CATHY~\cite{Wang13CATHY} uses a word co-occurrence network to compress the documents and performs topic discovery through an EM algorithm. \hide{Although it is designed for short, content-representative text, the word co-occurrence network modeling and the recursive clustering algorithm are applicable to long text as well.}
\item {STROD -- and its variations RTOD, RTOD\textsubscript{2}, RTOD\textsubscript{3}}. This is our scalable tensor recursive orthogonal decomposition method. We implement several versions to analyze our scalability improvement techniques: (i) RTOD: recursive tensor orthogonal decomposition without scalability improvement (Algorithm~\ref{alg:rtod}); (ii) RTOD\textsubscript{2}: RTOD plus the technique of avoiding creation of $M_2$; (iii) RTOD\textsubscript{3}: RTOD plus the technique of avoiding creation of $M_3$; and (iv) STROD: Algorithm~\ref{alg:strod} with the full scale-up technique.
\end{itemize}
We use an optimized Java implementation MALLET~\cite{MALLET}\hide{\footnote{\small\url{http://mallet.cs.umass.edu/}}} for the first three Gibbs sampling-based methods, and set the number of iterations to be 1000, which is the common practice. We implement CATHY and STROD in MATLAB because Java does not have good support with matrix computation and spectral algorithms.
\subsection{Scalability}
\begin{figure}
\centering
 \includegraphics[width=\linewidth]{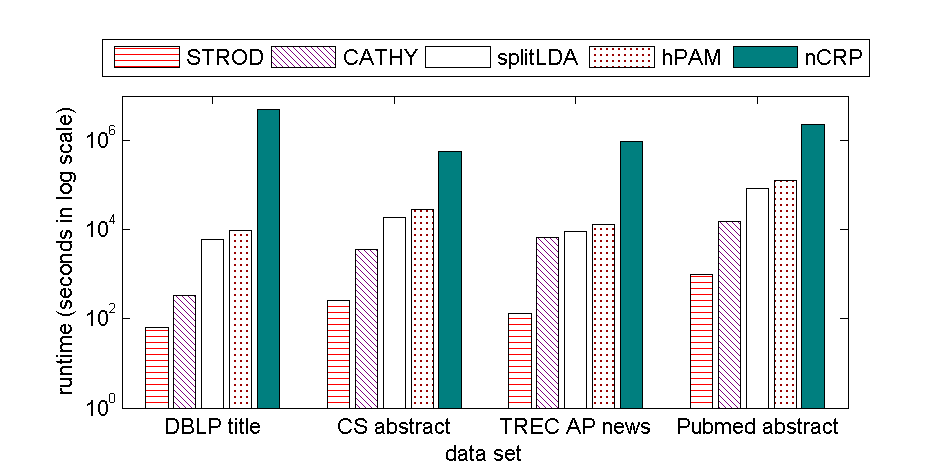}
\caption{\small Total runtime on each dataset, $H=2,C_t=5$}
\label{fig:time}
\end{figure}
\begin{figure}
 \centering
 \begin{subfigure}{.48\linewidth}
 \includegraphics[width=\linewidth]{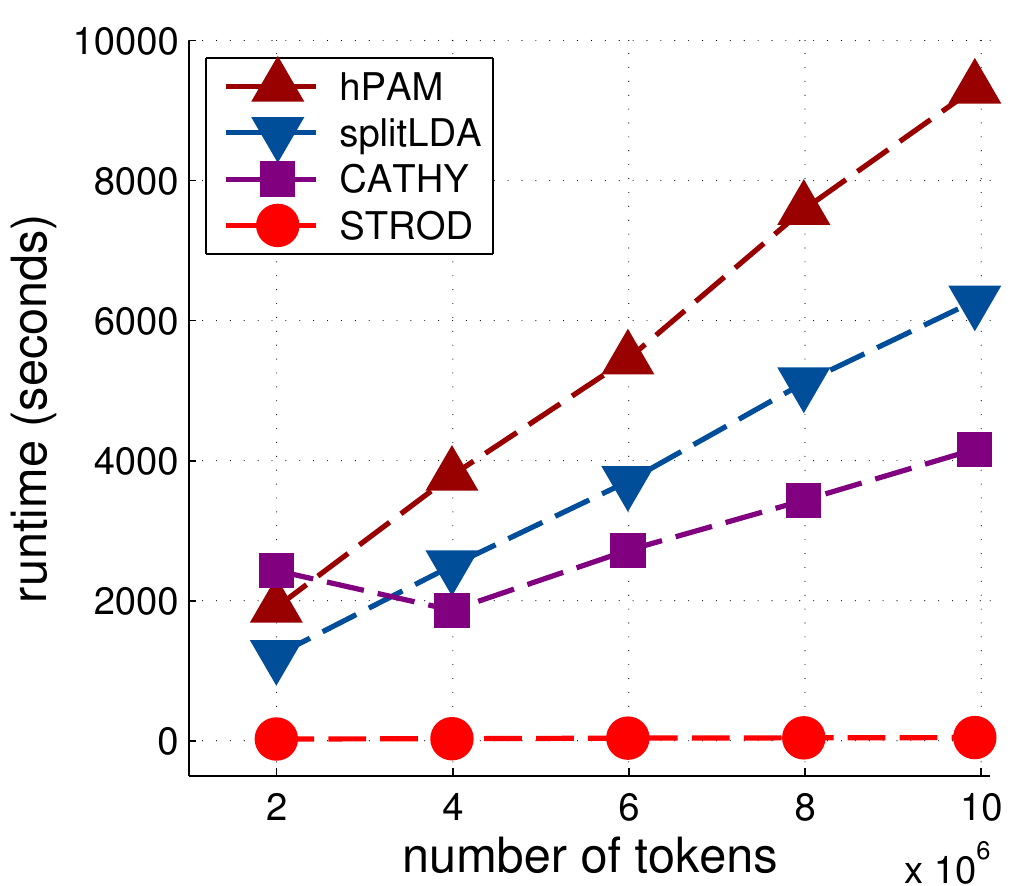}
 \caption{\small STROD is more efficient than the existing methods}
 \label{fig:sampledoc}
 \end{subfigure}
\begin{subfigure}{.48\linewidth}
 \centering
  \includegraphics[width=\linewidth]{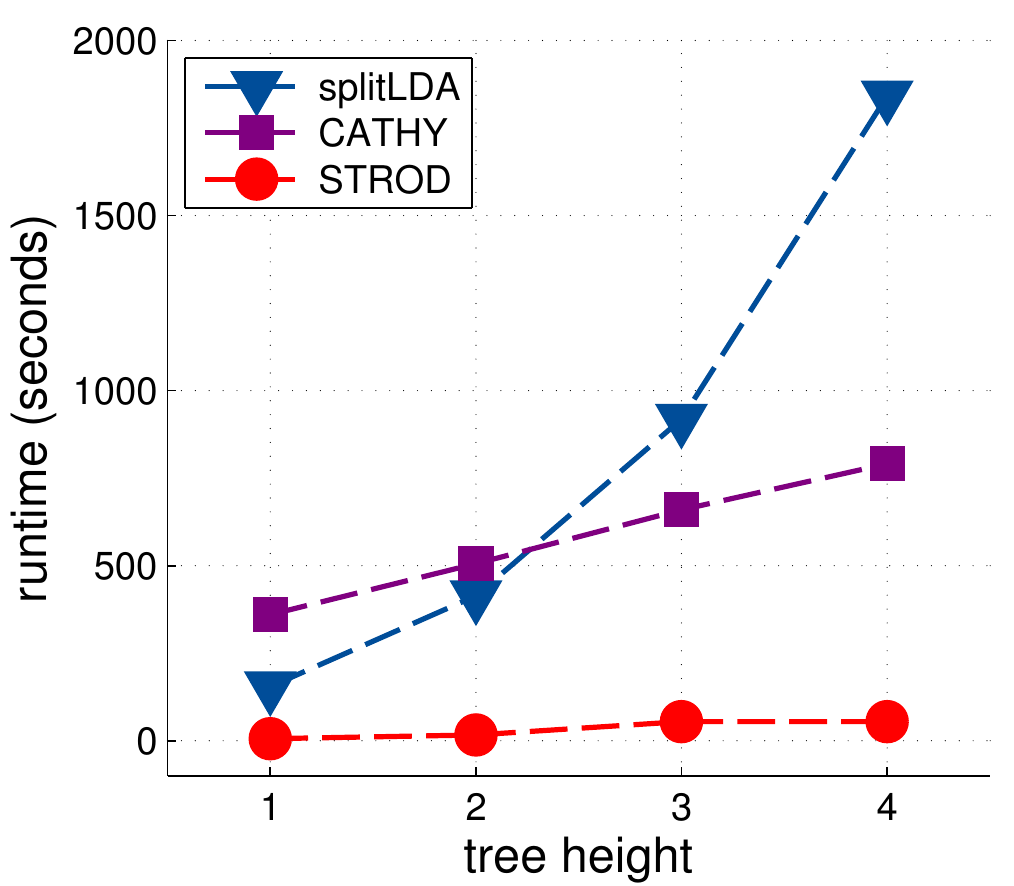}
\caption{\small Runtime w.r.t tree height}
 \label{fig:h}
\end{subfigure}
\begin{subfigure}{.48\linewidth}
 \centering
  \includegraphics[width=\linewidth]{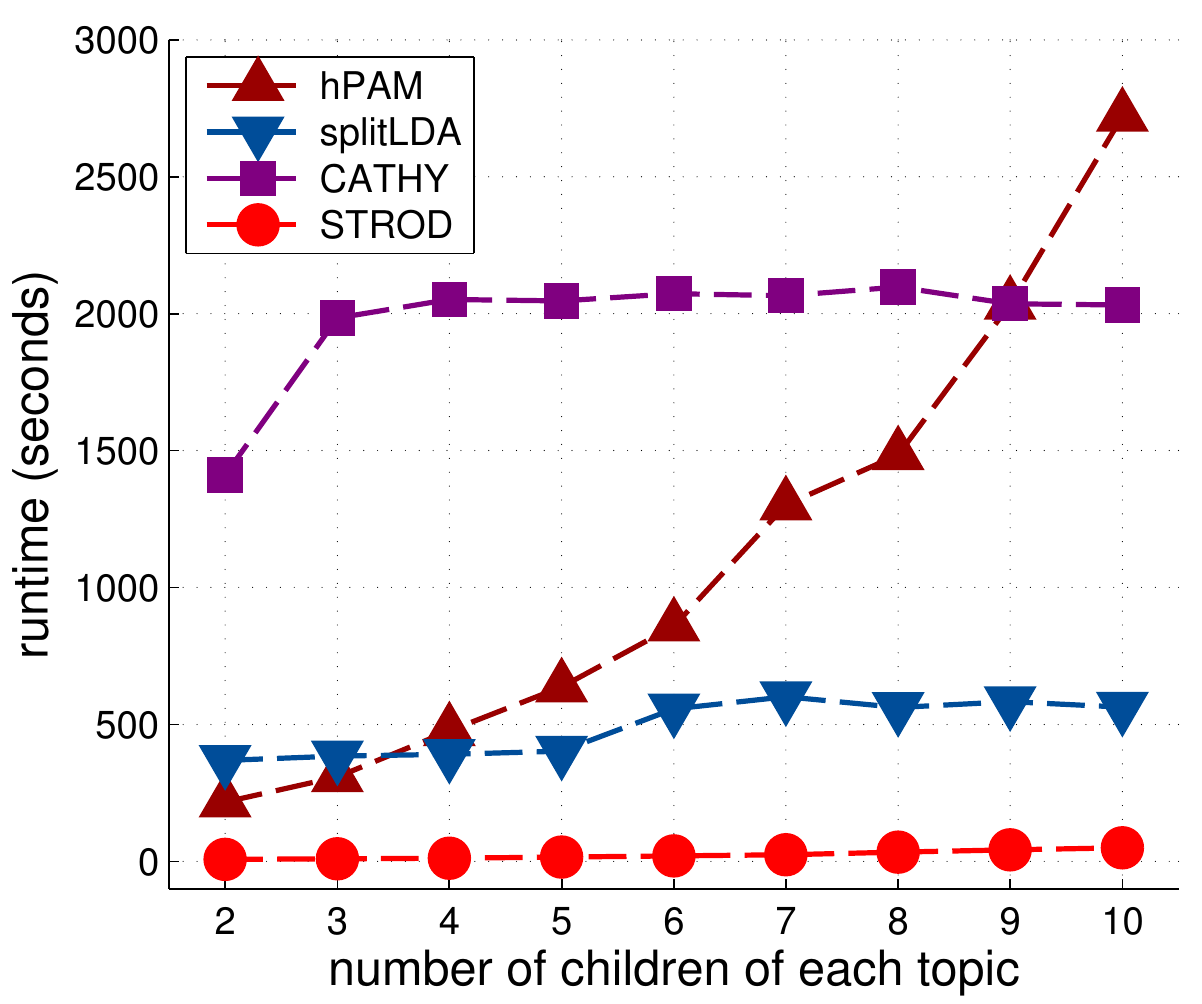}
\caption{\small Runtime w.r.t the number of children for each topic}
 \label{fig:ct}
\end{subfigure}
\begin{subfigure}{.48\linewidth}
 \includegraphics[width=\linewidth]{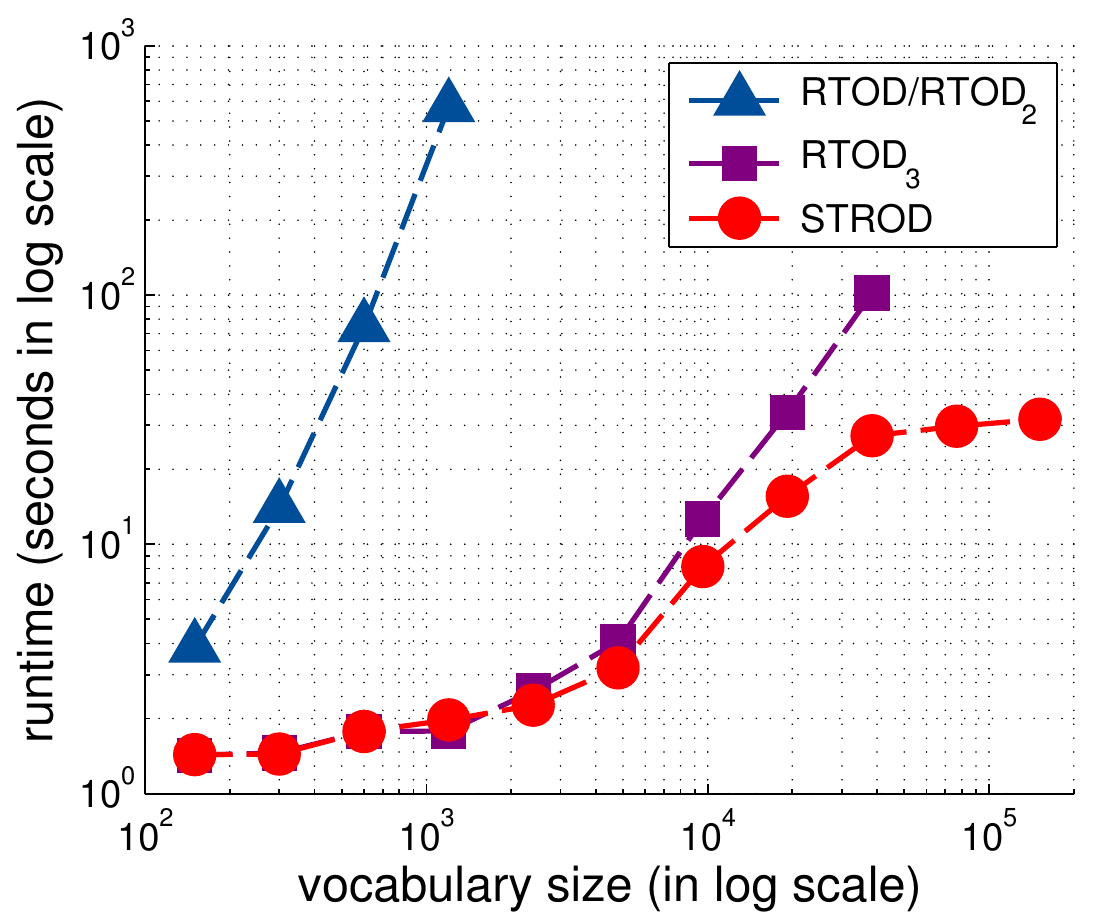}
 \caption{\scriptsize STROD scales much better than its variations (All except STROD fail to scale beyond 50K vocabulary size due to memory constraints)}
 \label{fig:rtod}
\end{subfigure}
\caption{Runtime varying with scale \hide{(DBLP title)}}
\label{fig:scale}
\end{figure}

Our first evaluation assesses the scalability of different algorithms, which is one focal point of this paper.

Figure~\ref{fig:time} shows the overall runtime in these datasets. STROD is several orders of magnitude faster than the existing methods. On the largest dataset it reduces the runtime from one or more days to 18 minutes. CATHY is the second best method in short documents such as titles and abstracts because it compresses the documents into word co-occurrence networks. But it is still more than 100 times slower than STROD due to many rounds of EM iterations. splitLDA and hPAM rely on Gibbs sampling, and the former is faster because it recursively performs LDA, and considers fewer dependencies in sampling. nCRP is two orders of magnitude slower due to its nonparametric nature.

We then conduct analytical study of the runtime growth with respect to different factors.
Figures~\ref{fig:sampledoc}-\ref{fig:ct} show the runtime varying with the number of tokens, the tree height and the tree width. We can see that the runtime of STROD grows slowly, and it has the best performance in all occasions. In Figure~\ref{fig:h},
\hide{Second, we vary the height $H$ of the tree. }we exclude hPAM because it is designed for $H=2$.
\hide{Third, we vary the number of children $C_t$ for each node in the tree. }In Figure~\ref{fig:ct}, we use the same number of child topics $C_t$ for each node for all the methods. 
We exclude nCRP from all these experiments because it takes too long time to finish ($>$90 hours with 600K tokens).

Figure~\ref{fig:rtod} shows the performance in comparison with the slower variations of STROD. Both RTOD and RTOD\textsubscript{2} fail to finish when the vocabulary size grows beyond 1K, because the third-order moment tensor $M_3$ requires $O(V^3)$ space to create. RTOD\textsubscript{3} also has limited scalability because the second order moment tensor $M_2\in\real^{V\times V}$ is dense. STROD scales up easily by avoiding explicit creation of these tensors.

\subsection{Robustness}

Our second evaluation assesses the robustness of different algorithms.
For each dataset, we sample 10, 000 documents and run each algorithm 10 times and measure the \emph{variance} among the 10 runs for the same method as follows. Each pair of algorithm runs generate the same number of topics, but their correspondence is unknown. For example, the topic $o/1$ in the first run may be close to $o/3$ in the second run. We measure the KL divergence between all pairs of topics between the two runs, build a bipartite graph using the negative KL divergence as the edge weight, and then use a maximum matching algorithm to determine the best correspondence  (top-down recursively). Then we average the KL divergence between matched pairs as the difference between the two algorithm runs.  Finally, we average the difference between all $10\times 9=90$ ordered pairs of algorithm runs as the final variance.
We exclude nCRP in this section, since even the number of topics is not a constant after each run. Due to space limitation, we report the variance on the first three datasets.

Table~\ref{tbl:robust} summarizes the results: STROD has lowest variance in all the three datasets. The other three methods based on Gibbs sampling have variance larger than 1 in all datasets, which implies that the topics generated across multiple algorithm runs are considerably different.

\begin{figure}
 \centering
 \begin{subfigure}{0.48\linewidth}
 \includegraphics[width=\linewidth]{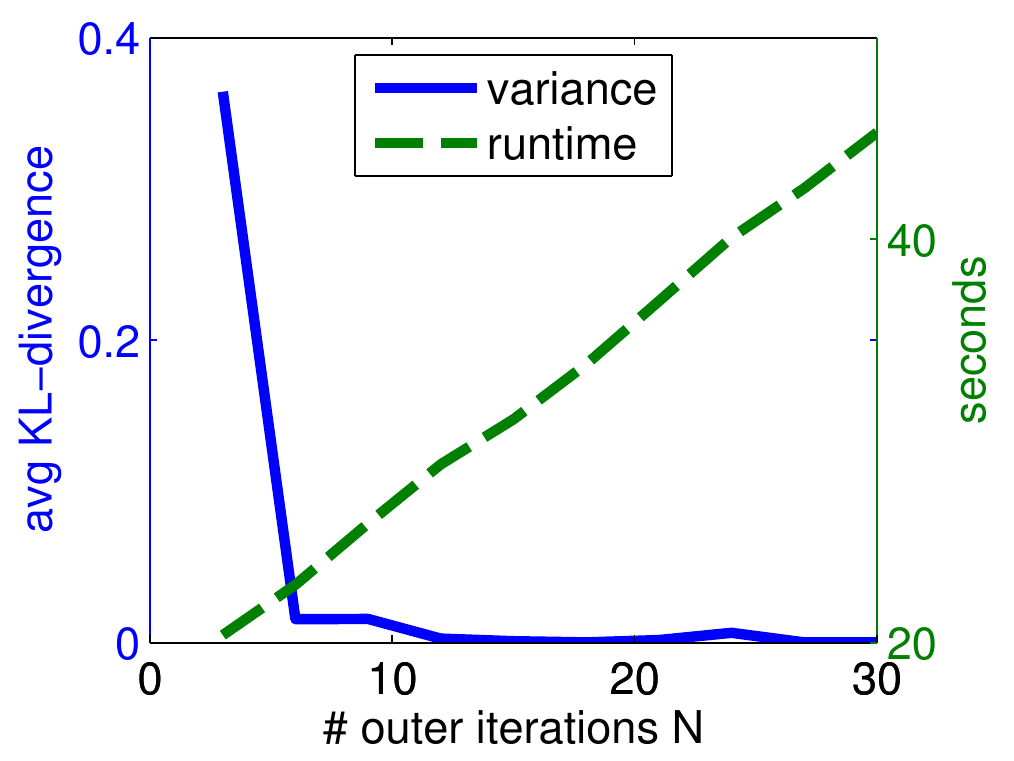}	
\caption{\small varying $N$ when $n=30$}
 \label{fig:outer}
 \end{subfigure}
\begin{subfigure}{0.48\linewidth}
 \includegraphics[width=\linewidth]{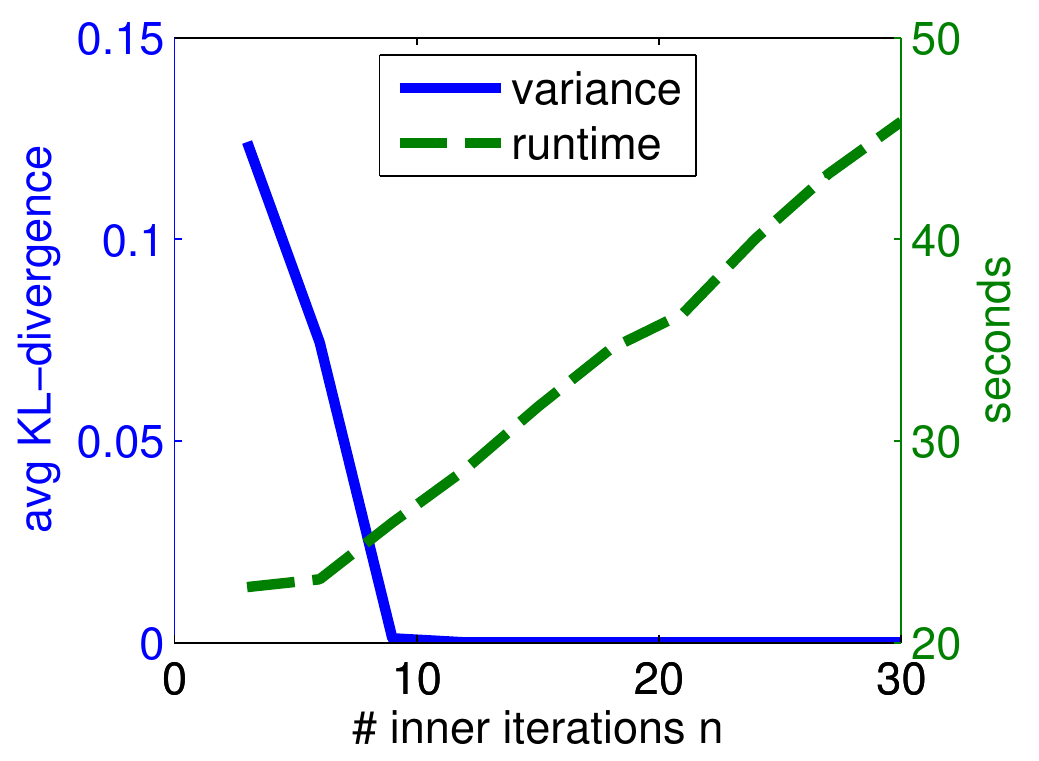}	
\caption{\small varying $n$ when $N=30$}
 \label{fig:inner}
\end{subfigure}
\caption{The variance and runtime {of STROD} when varying \# outer and inner iterations $N$ and $n$ (CS abstract)}
\label{fig:iter}
\end{figure}

We also evaluate the variance of STROD when we vary the number of outer and inner iterations $N$ and $n$. As shown in Figure~\ref{fig:iter}, the variance of STROD quickly diminishes when the number of outer and inner iterations grow to 10. The same trend is true for other datasets. This validates the theoretical analysis of their fast convergence and the guarantee of robustness.

In conclusion, STROD achieves robust performance with small runtime. It is stable and reliable to be used as a hierarchy construction method for large text collections.

\begin{table}
 \centering
\caption{\small The variance of multiple algorithm runs in each dataset}
\scriptsize
\begin{tabular}{ l  c  c  c  }
\toprule
{\textbf{Method}} & {\textbf{DBLP title}} & {\textbf{CS abstract}} & \textbf{TREC AP news} \\
\midrule
\textbf{hPAM} 					& 5.578 			& 5.715 		& 5.890 	\\
\textbf{splitLDA}	 & 3.393 			& 1.600 		& 	1.578		\\
\textbf{CATHY} 				& 17.34 			& 1.956 		& {1.418} 		\\
\textbf{STROD} 			& \textbf{0.6114} & \textbf{0.0001384} & \textbf{0.004522}  \\
\bottomrule
\end{tabular}
\label{tbl:robust}
\end{table}

\subsection{Interpretability}
Our final evaluation assesses the interpretability of the constructed topical hierarchy, via human judgment. We evaluate hierarchies constructed from DBLP titles and TREC AP news. For simplicity, we set the number of subtopics to be 5 for all topics. For hPAM, we post-process them to obtain the 5 strongest subtopics for each topic. For all the methods we use the same phrase mining and ranking procedure to enhance the interpretability. We do not include nCRP in this study because hPAM has been shown to have superior performance of it~\cite{Mimno07}.

In order to evaluate the topic coherence and parent-child relationship, we use two \emph{intrusion detection} tasks which were proposed in~\cite{Wang13CATHY} (adopting the idea in~\cite{Chang09}):

\begin{itemize}
\parskip -0.2ex
\item {Phrase Intrusion (PI)}: $X$ phrases are shown to an evaluator.  One is a top-10 phrase from a sibling topic, and the remaining ones come from the top-10 phrases of the same topic.
     Evaluators are asked to select the intruder phrase{, or to indicate that they are unable to make a choice}.
\item {Topic Intrusion (TI)}:  Evaluators are shown a parent topic $t$ and $X$ candidate child topics. $X-1$ of the child topics are actual children of $t$ in the generated hierarchy, and the remaining child topic is not. Each topic is represented by its top-5 ranked phrases. Evaluators are asked to select the intruder child topic{, or to indicate that they are unable to make a choice}.
\end{itemize}

For this study we set $X=4$. 160 Topic Intrusion questions and 200 Phrase Intrusion questions are randomly generated from the hierarchies constructed by these methods. We then calculate the agreement of the human choices with the actual hierarchical structure constructed by the various methods. We consider a higher match between a given hierarchy and human judgment to imply a higher quality hierarchy. For each method, we report the F1 measure of the questions answered `correctly' (matching the method) and consistently by three human judgers with CS background.

\begin{figure}[htp]
\centering
\includegraphics[width=\linewidth]{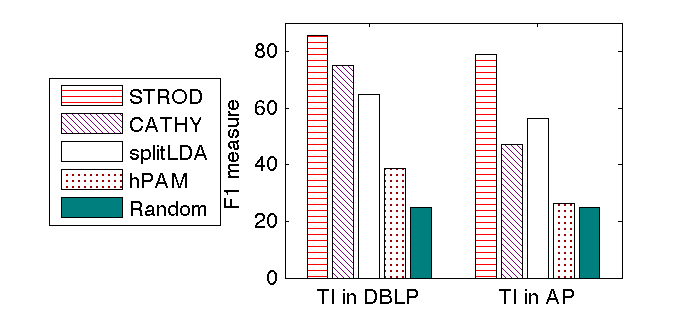}
\caption{\small Phrase intrustion and topic intrustion study}	
\label{fig:survey}
\end{figure}

Figure~\ref{fig:survey} summarizes the results. STROD is among the best performing methods in both tasks. This suggests that the quality of the hierarchy is not compromised by the strong scalability and robustness of STROD. As expected,  splitLDA and STROD perform similarly in PI task, since they share the same LDA process for one level. However, 
STROD has a more principled model and theoretically guaranteed inference method to construct the hierarchy. That lead to more meaningful parent-child relations, and thus better performace in TI task.

\begin{figure*}
\vspace{-0.2cm}
 \centering
\includegraphics[width=\textwidth]{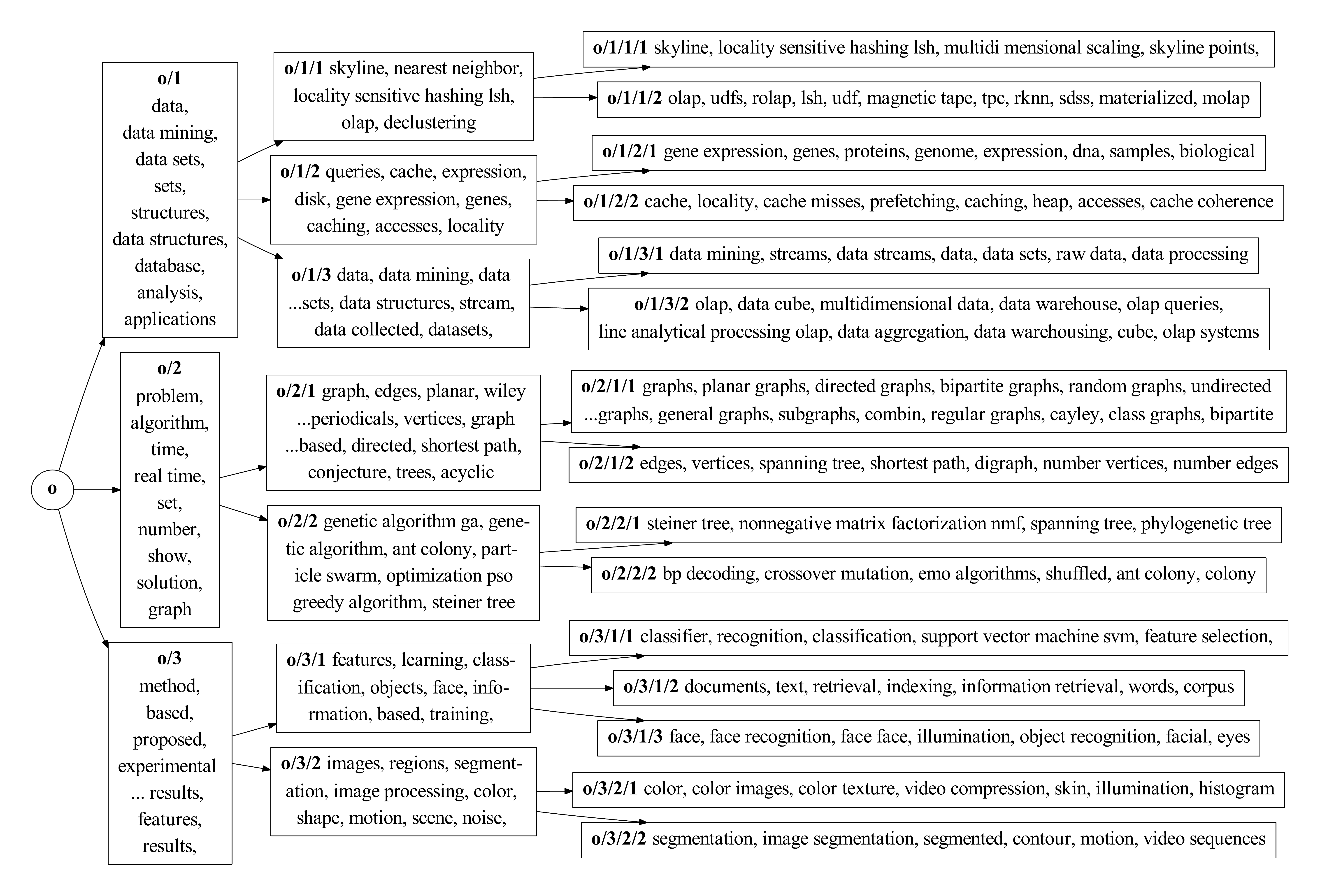}
\caption{\small Sample of hierarchy generated by STROD (two phrases only differing in plural/single forms are shown only once)\hide{For each method, we show the subtrees rooted at Level 2 that are the most likely to represent the topics of Information Retrieval and Databases. The ordering of words in each phrase are determined by the most frequent ordering in the documents, and }}
 \label{fig:hiertop}
\vspace{-0.1cm}
\end{figure*}

A subset of the hierarchy constructed from CS abstract is presented in Figure~\ref{fig:hiertop}. For each non-root node, we show the top ranked phrases. Node o/1 is about `data', while its children involves database, data mining and bioinformatics. The lower the level is, the more pure the topic is, and the more multigrams emerge ahead of unigrams in general. 

\section{Conclusions}\label{sec:conc}

In this work, we tackle the scalability and robustness challenge of topical hierarchy construction from large-scale text data. We design a novel framework to build the hierarchy recursively. 
A nice property of our hierarchical topic model permits dividing the inference problem into smaller subproblems. For robust inference, we leverage a theoretically promising tensor orthogonal decomposition technique. Utilizing the special structure of the tensor in our task, we dramatically overhaul the standard computing procedure to scale up the algorithm. By evaluating our approach on a variety of datasets, we demonstrate a huge computational advantage. Our algorithm generates stable and high-quality topic hierarchy 100-1000 times faster than the state-of-the-art, and the margin grows when the corpus size increases.

Our invention opens up numerous possibilities for future work.  On the application side, new systems can be built to support explorative data analysis in multiple granularity, domain knowledge learning, and OLAP in a large scale.  On the methodology side, the advantage of STROD can be further fulfilled by parallelization and adaptation to dynamic text collections\hide{ or using more advanced spectral decomposition methods}.  We would also like to study how to apply or extend the powerful STROD technique to solve other problems in big data analysis, such as mining latent entity structures in heterogeneous information networks.


\bibliographystyle{abbrv}
\scriptsize
\bibliography{SCATHY}


\subsection*{Appendix A. Discussion of Theorem~\ref{the:unique}}
Theorem~\ref{the:unique} relies on several non-trivial claims: i) the orthogonal decomposition of $\widetilde{T}$ is unique; ii) the power iteration converges robustly and quickly to the eigenpair; and iii) a pair of $(\widetilde{\lambda_z},\widetilde{v_z})$ uniquely determines a pair of $(\lambda_z,v_z)$. The first two are proved in Anandkumar \etal~\cite{anandkumar2012}, and the third can be proved by a similar proof of Theorem 4.1 in Anandkumar \etal~\cite{anandkumar12}. We omit the details here. To see why these claims are non-trivial, we notice that the decomposition of $M_2= \sum_{z=1}^k \lambda_z v_z \otimes v_z $ is not unique. If $(\sigma_z,\mu_z)$ are orthonormal eigenpairs of $M_2$, then for any orthonormal matrix $O\in\real^{k\times k}$, $M_2=\sum_{z=1}^k \sigma_z (O\mu_z)\otimes (O\mu_z)$. So there are infinite number of ways of decomposition if we only consider second order moments. This explains why CATHY's word co-occurrence network model has no robust inference method, since the word co-occurrence information is equivalent to the second order moments.

Anandkumar \etal~\cite{anandkumar2012} also provides perturbation analysis about Algorithm~\ref{alg:tod}. When $N$ and $n$ are sufficiently large, the decomposition error is bounded by the error $\epsilon$ of empirical moments from theoretical moments. The number of required inner loop iterations $n$ grows in a logarithm rate with $k$, and the outer loop $N$ in a polynomial rate. They also proposed possible stopping criterion to reduce the number of trials of the random restart. Since the number of components $k$ is bounded by a small constant $K\approx 10$ in our task, the power iteration update is very efficient, and we observe that $N=n=30$ are sufficient.

\subsection*{Appendix B. Hyperparameter Learning}\label{sec:hyper}
\noindent
\textbf{1. Selection of the number of topics.}
We discuss how to select $C_t$ when the tree width $K$ is given\hide{ but the number of child topics of each node is not}.  We first compute the largest $K$ eigenvalues of $E_2$ in Line~\ref{line:spectral}, and then select the smallest $k$ such that the first $k$ eigenpairs form a subspace that is good approximation of $E_2$'s column space\hide{the summation of the first $k$ eigenvalues $\sigma_1,\dots,\sigma_k$ is larger than a percentage threshold of the summation of all the $K$ eigenvalues}. This is similar to the idea of using Principle Component Analysis (PCA) to select a small subset of the eigenvectors as basis vectors. The \emph{cumulative energy} $g(k)$ for the first $k$ eigenvectors is defined to be $g(k)=\sum_{z=1}^k \sigma_z$. And we choose the smallest value of $k$ such that $\frac{g(k)}{g(K)}>\eta$, and let $C_t=k$. $\eta\in[0,1]$ controls the required energy of the first $k$ eigenvectors, and can be tuned according to the application. When $\eta=1$ a full $K$-branch tree will be constructed. When $\eta=0$ the tree contains a single root node because $C_o=0$. Typically $\eta$ between 0.7 and 0.9 results in reasonable children numbers.

\smallskip \noindent
\textbf{2. Learning Dirichlet prior.} First, we note that the individual prior $\alpha_{t,z}$ can be learned by the decomposition algorithm, when the summation $\alpha_{t,0}$ of $\alpha_{t,1}$ to $\alpha_{t,C_t}$\hide{ \from{JH}{Pls check! correct?}} is given to perform the inference for topic $t$. This already largely reduces the number of hyperparameters that are needed to be given. Large $\alpha_{t,0}$ indicates that $t$'s subtopics tend to be mixed together in a document, while small $\alpha_{t,0}$ suggests that a document usually talks about only a few of the subtopics. When $\alpha_{t,0}$ approaches 0, one expects a document to have only one subtopic of $t$. So $\alpha_{t,0}$ can usually be set empirically according to the prior knowledge of the documents, such as 1 to 100\hide{, without affecting results very sensitively}.

If one wants to learn $\alpha_{t,0}$ automatically, we propose a heuristic method hereby. Suppose the data are generated by an authentic $\alpha^*_{t,0}$, and the moments are computed using the same $\alpha^*_{t,0}$, then the decomposition result should satisfy $\sum_{z=1}^{C_t}\alpha_{t,z}=\alpha_{t,0}$ exactly. However, if one uses a different $\alpha_{t,0}$ to compute the moments, the moments could deviate from the true value and result in mismatched $\alpha_{t,z}$. The discrepancy between returned $\sum_{z=1}^{C_t}\alpha_{t,z}$ and initial $\alpha_{t,0}$ indicate how much $\alpha_{t,0}$ deviates from the authentic value. \hide{Moreover, if $\alpha_{t,0}>\alpha^*_{t,0}$, the empirical $M_2$ is smaller than the authentic $M_2$, and $\sum_{z=1}^{C_t}\alpha_{t,z}<\alpha^*_{t,0}$. }So we can use the following fixed-point method to learn $\alpha_{t,0}$, where $\delta$ is learning rate. 
\begin{enumerate}
	\item Initialize $\alpha_{t,0}=1$;
	\item While (not converged)
	\begin{enumerate}
		\item Perform tensor decomposition for topic $t$ to update $\alpha_{t,z},z=1,\ldots,C_t$ ;
		\item $\alpha'_{t,0}=\sum_{z=1}^{C_t}\alpha_{t,z}$;
		\item Update $\alpha_{t,0}\leftarrow\alpha_{t,0}+ \delta(\alpha'_{t,0}-\alpha_{t,0})$;
	\end{enumerate}
\end{enumerate}

\end{document}